%% file: main.tex
\documentclass[10pt,twocolumn,letterpaper]{article}

\usepackage[pagenumbers]{wacv} 


\usepackage{tabularx}
\usepackage{booktabs}
\usepackage{placeins}
\usepackage{dblfloatfix}
\definecolor{wacvblue}{rgb}{0.21,0.49,0.74}
\usepackage[pagebackref,breaklinks,colorlinks,allcolors=wacvblue]{hyperref}

\usepackage{graphicx}
\usepackage{booktabs}
\usepackage{amsmath,amssymb}
\usepackage[ruled,vlined,linesnumbered]{algorithm2e}
\usepackage{xspace}
\usepackage{adjustbox}
\usepackage{caption}
\newcommand{\dmodel}{d_{\text{model}}}
\newcommand{\dsae}{d_{\text{sae}}}

\def\confName{WACV}
\def\confYear{2027}

\title{Dr.~No: Feature Steering for Abstention in Vision-Language Models}

\author{
Farid Hasanov$^{1}$ \qquad Riccardo Renzulli$^{2}$ \qquad Bahram Zonooz$^{1}$\\
Mykola Pechenizkiy$^{1}$ \qquad Decebal Constantin Mocanu$^{3}$\\[0.6em]
$^{1}$Eindhoven University of Technology \qquad $^{2}$University of Turin\\
$^{3}$University of Luxembourg\\[0.4em]
{\small Corresponding author: \texttt{f.e.o.hasanov@tue.nl}}
}

\begin{document}
\maketitle

\begin{abstract}
Large foundation models often answer even when the available visual or textual evidence is insufficient to support a reliable conclusion. This failure to abstain makes them hard to trust in high-stakes or information-seeking settings, where a confident but unsupported answer can be worse than no answer at all.  It also raises a natural question: can the internal mechanisms behind this behaviour be identified and controlled at inference time? We present Dr.~No, a feature-steering approach that mines sparse autoencoder (SAE) features associated with abstention and commitment, then modulates a small set of them during the prefill pass, without retraining the base model. 

Dr.~No is evaluated on LLaVA-1.6 Mistral 7B across multimodal abstention benchmarks, including UNK-VQA and MM-UPD, and additionally benchmarked on language reasoning tasks using Qwen3.5-2B on AbstentionBench subsets. Against multiple uncertainty-based abstention baselines, feature steering achieves a more favourable balance between correct answering and appropriate abstention than post-hoc rejection methods. The gains are strongest on UNK-VQA and on MM-UPD cases involving image--question or question--answer incompatibility, and transfer most clearly to mathematical reasoning subsets of AbstentionBench. However, performance is more limited on absent-answer detection and retrieval-heavy multi-hop QA, such as MuSiQue. These results suggest that sparse internal features provide lightweight control handles for shaping uncertainty-aware behaviour in foundation models, while also showing that abstention control is source-dependent rather than governed by a single universal signal.
\end{abstract}
\begin{figure}[t]
    \centering
    \includegraphics[width=1.00\columnwidth]{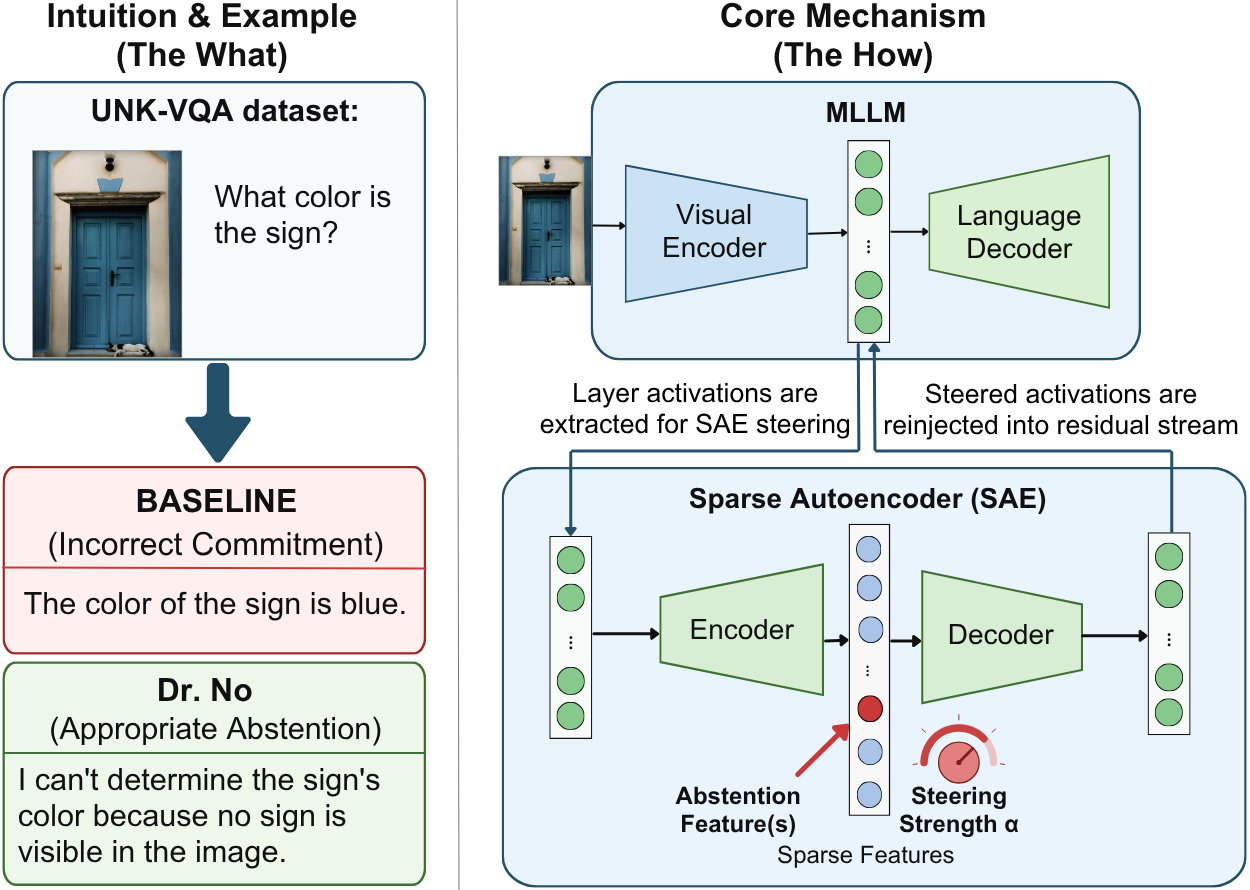}
    \caption{Given an ambiguous question, the baseline model commits to an answer, while Dr.~No steers SAE-derived features at inference time to promote appropriate refusal. The base model remains frozen.}
    \vspace{-1.6em}
    \label{fig:myimage}
\end{figure}
\section{Introduction}

Large foundation models are increasingly deployed in settings where reliable behaviour requires more than producing plausible answers. Across multimodal and language-only tasks, queries may be unanswerable because the required visual or textual evidence is missing, inconsistent, or insufficient~\cite{unkvqa,mmupd,abstentionbench}. In such cases, a model should abstain rather than guess. However, standard training and evaluation pipelines typically reward answer production, creating a bias toward committing even when the evidence does not warrant a reliable conclusion~\cite{arrogance}. This leaves models prone to overconfident responses when the evidence does not support a reliable answer~\cite{pope,kadavath2022language,guo2017}.

Most existing abstention methods address this problem after generation. They estimate uncertainty from output-level signals such as maximum softmax probability, entropy, sequence log-probability, self-evaluation, verbalized confidence, or disagreement across sampled responses~\cite{msp,malinin,kadavath2022language,canllmsexpress,kuhn2023semantic}. These methods are attractive because they are broadly applicable and require no retraining. However, they typically control abstention by thresholding a scalar uncertainty score, which can make the trade-off between refusing unanswerable queries and preserving correct answers hard to control.

This paper explores a complementary approach: controlling abstention through sparse internal features. Sparse autoencoders decompose residual-stream activations into sparse feature activations, some of which correspond to interpretable semantic or behavioural directions. We hypothesise that abstention and answer commitment are partially represented by such features, and that modulating a small number of them can shift a model's willingness to answer before the final output is produced.

We introduce \emph{Dr.~No}, an inference-time feature-steering method for abstention in vision-language models. Dr.~No mines SAE features that discriminate between abstention and commitment on a separate corpus, then intervenes on residual-stream activations during the prefill pass by increasing abstention-associated features and/or suppressing commitment-associated features. We also study gated variants, where steering is applied only when abstention-related features are active, reducing the risk of globally pushing the model toward refusal. The method is lightweight, model-internal, and compatible with frozen base models. Figure~\ref{fig:myimage} illustrates our approach: where the baseline commits to an answer on an ambiguous query, Dr.~No steers SAE-derived abstention/commitment features at inference time to promote appropriate refusal while leaving the base model frozen.

We evaluate Dr.~No on LLaVA~1.6 Mistral 7B \cite{llava} using two multimodal abstention benchmarks, UNK-VQA \cite{unkvqa} and MM-UPD \cite{mmupd}, which test whether models can distinguish answerable visual questions from cases where the image, question, or answer set does not support a reliable response. MM-UPD is particularly useful because it pairs solvable and unsolvable variants, allowing us to measure whether a method preserves correct answering while improving abstention. To assess our approach beyond vision-language inputs, we also evaluate Qwen3.5-2B \cite{qwen3.5} on AbstentionBench \cite{abstentionbench}, focusing on mathematical reasoning and multi-hop question answering. Our experiments show that SAE-based steering can improve the balance between answering and abstaining, while also revealing clear limits. On UNK-VQA, Dr.~No outperforms uncertainty-based rejection baselines in harmonic mean, indicating a more favourable trade-off between answering solvable questions and abstaining on unanswerable ones. On MM-UPD, the gains are source-dependent: steering is most effective for image--question and question--answer incompatibilities, whereas uncertainty-based rejection remains competitive when the correct answer is absent from the candidate set. In the language setting, steering transfers most clearly to mathematical reasoning tasks, but is less effective on MuSiQue \cite{trivedi2022musique}, where abstention depends more heavily on evidence retrieval and multi-hop composition. These findings suggest that sparse features can provide useful abstention-control directions, but different sources of unanswerability may require different control signals.

\paragraph{Contributions.}
This paper makes the following contributions:
\begin{itemize}
\item We propose Dr.~No, an inference-time feature-steering method that uses SAE-derived sparse features to modulate abstention behaviour in frozen foundation models without retraining.
\item We introduce a feature-mining and intervention procedure that identifies abstention- and commitment-associated SAE features from a separate mining corpus and uses them to steer the model's prefill-stage activations.
\item We evaluate Dr.~No on LLaVA~1.6 Mistral 7B across UNK-VQA and MM-UPD, showing that feature steering can improve the answer--abstain trade-off relative to post-hoc uncertainty baselines, with gains that depend on the source of unanswerability.
\item We extend the evaluation beyond vision-language to a language-only benchmark, AbstentionBench with Qwen3.5-2B, where SUM-mined \cite{sum} features prove most effective on mathematical reasoning tasks while exposing limitations on retrieval-heavy multi-hop QA.
\end{itemize}

We will release the Dr.~No feature-mining, steering, and evaluation code to support reproducibility.

\section{Background and Related Work}

Reliable foundation models must answer when sufficient evidence is available and abstain when the available visual or textual evidence is inadequate for drawing a reliable conclusion. This framing aligns with selective prediction, where a useful model should preserve coverage on answerable inputs while refusing cases that lack adequate evidential support \cite{selective1,selective2,selective3}. When reasoning over images and text, unanswerability can arise from diverse sources; illustrative cases include absent queried objects, incompatibility between the image and question, and answer sets in which no option is valid \cite{unkvqa,mmupd}. Text-only abstention benchmarks similarly probe false premises, underspecification, outdated information, and other cases in which the provided context is insufficient to justify an answer \cite{abstentionbench}. Since standard training often rewards answer production, models can remain biased toward confident commitment even when the evidence is insufficient \cite{arrogance,pope,kadavath2022language}.
\paragraph{Uncertainty-based abstention.}
Post-hoc abstention methods differ mainly in where they obtain their rejection signal. Distributional methods use token probabilities, such as maximum softmax probability, predictive entropy, or sequence log-probability, as confidence proxies \cite{msp,malinin}. Self-assessment methods elicit a second-pass judgement from the model, either through calibrated probabilities such as $P(\mathrm{True})$ or through verbalized confidence scores \cite{kadavath2022language,canllmsexpress}. Sampling-based methods estimate uncertainty from variation across multiple generations, with semantic entropy clustering responses by meaning before computing disagreement \cite{kuhn2023semantic}. Dr.~No is complementary to these approaches: rather than rejecting an already-produced answer, it intervenes on internal activations before the answer is generated.

\paragraph{Internal representations and activation steering.}
A complementary line of work controls model behaviour by modifying internal
activations at inference time. Representation engineering and activation-steering
methods derive directions from contrastive examples, then add or subtract them
during generation to influence behaviours such as sentiment, harmlessness, and
safety refusal~\cite{zou2023representation,turner2023steering}; for instance, ITI
shifts activations along a learned truthfulness direction at a small number of
attention heads~\cite{inference_eliciting}, and safety refusal can be ablated or induced through
a single residual-stream direction~\cite{safety_refusal}. Such interventions
are relevant to abstention because they act before the final answer is produced.
Among them, safety refusal is the closest analogue, but it is driven by a
different cause: there the model declines because a request is harmful, whereas
epistemic abstention is triggered when the available evidence is insufficient.
Dr.~No asks whether similar internal directions can control this distinct
boundary, between answer commitment and evidence-sensitive abstention.

\paragraph{Sparse autoencoders and multimodal feature steering.}
Sparse autoencoders decompose dense hidden states into sparse learned features, offering more interpretable activation directions in language models~\cite{huben,bricken2023monosemanticity,gao}. Recent work extends this approach to vision-language models, showing that SAE features can capture visual and cross-modal concepts and support causal interventions~\cite{sae_pach}. The closest work to ours is SAVE, which uses SAE features linked to visual information processing to mitigate object hallucination~\cite{save}. More broadly, Dr.~No builds on the insight that SAE-derived features can provide lightweight control handles for frozen multimodal models, but differs in objective and evaluation target: rather than steering object-presence features, we mine abstention- and commitment-associated features to control whether the model should answer at all, framing the problem as a selective-prediction trade-off between refusing unsupported inputs and preserving correct answers on supported ones. On the unanswerable-VQA side specifically, CLIP-UP~\cite{clipup} also targets this problem, but injects an external CLIP-derived image--text compatibility signal into a frozen VLM via additional learned components, whereas Dr.~No steers SAE-aligned residual-stream activations already present in the target model, without external cross-modal signals or added components. This also bounds CLIP-UP's scope, since it has no analogue in language-only settings, where Dr.~No is further evaluated (Table~\ref{tab:qwen3_abstentionbench}).
\section{Method}
\label{sec:method}

Dr.~No is an inference-time feature-steering method for improving abstention behaviour in frozen foundation models. The method uses SAE features as lightweight control directions: it first identifies sparse features associated with abstention and answer commitment directions, then modulates the corresponding SAE decoder directions during the prefill pass of generation. Neither the base model nor the SAE is retrained.

The method has three components. First, we build labelled mining corpora containing inputs for which the model should either answer or abstain. Second, we use these corpora to identify SAE features whose activations distinguish abstention-oriented states from commitment-oriented states. Third, at inference time, we add a steering vector to the model's residual stream, either uniformly or through an input-dependent gate that applies stronger steering when abstention-related features are already active.

\subsection{Overview}
\label{sec:method_overview}

Let $f_\theta$ denote a frozen transformer model and let an SAE be trained on residual-stream activations at layer $\ell$. For an input $x$, we extract the residual activation at the decision position, encode it into sparse SAE feature activations, and use selected SAE decoder directions as intervention directions.

Dr.~No assumes access to a labelled mining set
\[
\mathcal{D}_{\mathrm{mine}}
=
\{(x_i, y_i)\}_{i=1}^{N},
\]
where $y_i \in \{\mathrm{abstain}, \mathrm{commit}\}$. The mining set is used only to identify intervention features and calibrate optional gates. Test examples from the downstream evaluation benchmarks are not used for feature mining.

For each target layer, the method produces two ranked feature sets:
\[
\mathcal{A}
\quad \text{and} \quad
\mathcal{C},
\]
where $\mathcal{A}$ contains abstention-associated features and $\mathcal{C}$ contains commitment-associated features. At inference time, Dr.~No increases activation along abstention-associated directions, suppresses commitment-associated directions, or combines both interventions.

\subsection{Mining corpora}
\label{sec:mining_corpora}

\paragraph{Multimodal mining corpus.}
For the multimodal setting, we construct a controlled mining corpus from the GQA balanced training split~\cite{gqa}. Each example is formatted as a multiple-choice visual question with an explicit \emph{Unanswerable} option. For answerable examples, the gold answer is placed in a randomly selected answer slot and the target label is \emph{commit}. Unanswerable examples are constructed from four sources. \emph{Semantic drift} pairs a question with a different image, creating a mismatch between the question and visual context. \emph{Object absence} examples use scene-graph annotations to ask about objects that are not present in the image. \emph{False premise} examples introduce visually unsupported assumptions into the question. \emph{Answer dropping} keeps the image and question valid but removes the correct answer from the candidate option set.

Although our experiments instantiate this corpus with image-question pairs, the construction only requires a context, a question, candidate answers, and an abstention option. It is therefore not specific to GQA: analogous unanswerable mining corpora can be built from other datasets by applying suitable transformations to their contexts.

\paragraph{Text mining corpus.}
For the text-only setting, we use Synthetic Unanswerable Math (SUM) as the feature-mining corpus \cite{sum}. SUM provides an existing synthetic abstention dataset of mathematical word problems with binary abstain/commit labels, so unlike the multimodal setting, we do not need to construct new unanswerable examples through image--question or answer-option perturbations.

\subsection{SAE feature mining}
\label{sec:feature_mining}

For each layer under consideration, we run the target model on the mining corpus and collect residual-stream activations at the location where the SAE was trained. Activations are extracted at the final decision position, corresponding to the point at which the model has processed the context, question, candidate answers, and abstention option.

Because the base model does not always abstain on synthetic unanswerable inputs, feature mining is performed after behavioural filtering. Abstention features are mined from unanswerable examples on which the unsteered model correctly selects \emph{Unanswerable}, and commitment features are mined from answerable examples on which it selects the correct answer. This links the selected SAE features to realized abstention and commitment states rather than to synthetic labels alone. We report the resulting sample counts, class balance, and filtering details in Appendix~\ref{app:implementation_details}. Given a residual activation $h^{(\ell)} \in \mathbb{R}^{d}$ at layer $\ell$, the SAE maps it to sparse feature activations
\begin{equation}
a(h^{(\ell)})
=
\mathrm{TopK}
\left(
\mathrm{ReLU}
\left(
W_{\mathrm{enc}} h^{(\ell)} + b_{\mathrm{enc}}
\right)
\right),
\end{equation}
where $W_{\mathrm{enc}}$ and $b_{\mathrm{enc}}$ are the SAE encoder parameters. The SAE reconstructs the activation as
\begin{equation}
\hat{h}^{(\ell)}
=
W_{\mathrm{dec}} a(h^{(\ell)}) + b_{\mathrm{dec}}
=
\sum_{j=1}^{F} a_j(h^{(\ell)}) d_j + b_{\mathrm{dec}},
\end{equation}
where $F$ is the number of SAE features and
\[
d_j = W_{\mathrm{dec}}[:,j]
\]
is the decoder direction associated with feature $j$.

For each SAE feature $j$, we measure how strongly its activation separates abstain-labelled examples from commit-labelled examples. We use two complementary scores. The first is the mean activation difference,
\begin{equation}
\Delta_j
=
\mathbb{E}_{x \sim \mathcal{D}_{\mathrm{abstain}}}
\left[
a_j(h^{(\ell)}(x))
\right]
-
\mathbb{E}_{x \sim \mathcal{D}_{\mathrm{commit}}}
\left[
a_j(h^{(\ell)}(x))
\right],
\end{equation}
which identifies features that are more active in abstention states. The second
is the single-feature area under the receiver operating characteristic curve (AUROC), obtained by using $a_j(h^{(\ell)}(x))$ as a scalar detector of the
abstain label.

Features with large positive $\Delta_j$ and high AUROC are treated as abstention-associated features. Features with large negative $\Delta_j$, or equivalently features that are more active on commit-labelled examples, are treated as commitment-associated features. We rank features under each statistic and retain features that are consistently highly ranked. This gives two ordered feature lists per layer:
\[
\mathcal{A}
=
(a_1, a_2, \ldots),
\qquad
\mathcal{C}
=
(c_1, c_2, \ldots).
\]
The top-$k$ features from each list are selected before downstream evaluation and are fixed for all benchmark examples.

\subsection{Feature steering}
\label{sec:steering}

Dr.~No intervenes by adding SAE decoder directions to the model's residual stream. Let $h^{(\ell^\star)}_t$ denote the residual-stream activation at target layer $\ell^\star$ and token position $t$. For each selected feature $j$, we use the unit-normalised decoder direction
\begin{equation}
\bar{d}_j
=
\frac{d_j}{\|d_j\|_2 + \epsilon},
\end{equation}
where $\epsilon$ is a small numerical-stability constant. Normalising decoder directions prevents features with larger decoder norms from receiving disproportionately large effective intervention strength.

For a target layer, let $\mathcal{A}_k$ and $\mathcal{C}_k$ denote the top-$k$ abstention-associated and commitment-associated features. We define aggregate abstention and commitment directions as
\begin{equation}
v_A
=
\sum_{j \in \mathcal{A}_k} \bar{d}_j,
\qquad
v_C
=
\sum_{j \in \mathcal{C}_k} \bar{d}_j.
\end{equation}

The final steering vector is
\begin{equation}
v(m,\alpha)
=
\alpha
\left(
s_A(m)v_A - s_C(m)v_C
\right),
\label{eq:steering_vector}
\end{equation}
where $\alpha$ is the maximum steering strength and $m$ denotes the steering mode. We consider three modes:

\begin{itemize}
    \item \textbf{Abstention-up} increases abstention-associated directions by setting $s_A(m)=1$ and $s_C(m)=0$.
    \item \textbf{Commitment-down} suppresses commitment-associated directions by setting $s_A(m)=0$ and $s_C(m)=1$.
    \item \textbf{Bidirectional} combines both effects by setting $s_A(m)=s_C(m)=1$.
\end{itemize}

At inference time, we modify the target-layer residual stream as
\begin{equation}
\tilde{h}^{(\ell^\star)}_t
=
h^{(\ell^\star)}_t
+
v(m,\alpha).
\label{eq:steering_apply}
\end{equation}
Our main configurations use a \emph{gated} version of this update. Instead of applying the full vector to every input, we scale it by an input-dependent gate $g(x)\in[0,1]$ computed from the model's own abstention-related SAE activations at the decision token,
\begin{equation}
\tilde{h}^{(\ell^\star)}_t
=
h^{(\ell^\star)}_t
+
g(x)\,v(m,\alpha),
\label{eq:gated_apply}
\end{equation}
so that steering is applied more strongly when the input already activates the mined abstention features and is suppressed on inputs that do not, limiting over-abstention on answerable examples. Setting $g(x)\equiv 1$ recovers the ungated update of Eq.~\eqref{eq:steering_apply}. The definition of $g(x)$ and its calibration are deferred to Appendix~\ref{app:gated_steering}.

The intervention is applied during the prefill pass over the prompt tokens. Generated-token activations are left unchanged. This design modifies the model state after it has processed the image, question, and candidate answers, while avoiding continual intervention throughout autoregressive decoding.

\subsection{Dictionary-free baselines}
\label{sec:dictionary_free}

To isolate the contribution of the sparse dictionary itself, we also implement two simpler, dictionary-free interventions. \emph{Logit bias} shifts the model's output distribution by adding a fixed scalar directly to the abstention option's logit before greedy decoding. \emph{Dense steering} applies the same steering-vector formulation as Dr.~No (Eq.~\eqref{eq:steering_vector}), but intervenes directly on raw residual-stream activations rather than on sparse SAE decoder directions. Both are tuned and evaluated under the same protocol as Dr.~No.

\section{Experimental Setup}
\label{sec:experimental_setup}

We evaluate whether SAE-based feature steering improves the trade-off between
answering and abstaining across multimodal and language abstention settings.
This section describes the evaluation benchmarks, models and intervention setup, baselines, metrics, and implementation details.

\begin{table}[t]
\centering
\caption{Final best Dr.~No steering configurations. Hyperparameters are selected before test evaluation and then fixed for all reported test examples.}
\label{tab:drno_config}
\small
\setlength{\tabcolsep}{3.5pt}
\begin{tabular}{lcccccc}
\toprule
\textbf{Model} & \textbf{Mining} & \textbf{Layer} & \textbf{$k$} & \textbf{$\alpha$} & \textbf{Mode} & \textbf{Gate} \\
\midrule
LLaVA & GQA & 16\,/\,24 & 3 & 2.0\,/\,5.0 & both & \checkmark \\
Qwen & SUM & 15 & 3 & 1.0 & both & \checkmark \\
\bottomrule
\end{tabular}
\vspace{-1.25em}
\end{table}

\begin{table*}[!t]
\centering
\caption{
Results on UNK-VQA and MM-UPD for LLaVA~1.6 Mistral 7B. \textbf{Ans}/\textbf{Unans} denote answerable and unanswerable accuracy, and \textbf{HM} denotes their harmonic mean. \textbf{Std}/\textbf{UPD} denote standard and Unsolvable Problem Detection (UPD) accuracy, and \textbf{Dual} denotes paired standard--UPD accuracy. All values are percentages.
}
\label{tab:merged_results}

\scriptsize
\setlength{\tabcolsep}{4pt}
\resizebox{\textwidth}{!}{%
\begin{tabular}{lcccccccccccc}
\toprule
&
\multicolumn{3}{c}{\textbf{UNK-VQA}} &
\multicolumn{3}{c}{\textbf{MM-UPD (AAD)}} &
\multicolumn{3}{c}{\textbf{MM-UPD (IASD)}} &
\multicolumn{3}{c}{\textbf{MM-UPD (IVQD)}} \\
\cmidrule(lr){2-4}
\cmidrule(lr){5-7}
\cmidrule(lr){8-10}
\cmidrule(lr){11-13}
\textbf{Method}
& \textbf{Ans} & \textbf{Unans} & \textbf{HM}
& \textbf{Std} & \textbf{Upd} & \textbf{Dual}
& \textbf{Std} & \textbf{Upd} & \textbf{Dual}
& \textbf{Std} & \textbf{Upd} & \textbf{Dual} \\
\midrule
\multicolumn{13}{l}{\textit{Baseline}} \\
No steering
& 37.5 & 33.2 & 35.2
& 81.1 & 32.3 & 24.6
& 78.8 & 65.3 & 37.2
& 76.3 & 78.1 & 53.4 \\

No steering + abstention prompt
& 6.1 & 93.8 & 11.5
& 75.7 & 56.1 & 37.7
& 73.8 & 87.2 & 50.5
& 63.0 & 97.1 & 52.3 \\

\midrule
\multicolumn{13}{l}{\textit{Uncertainty-based rejection}} \\
P(True)
& 30.4 & 48.5 & 37.3
& 59.2 & 77.7 & 38.7
& 73.6 & 82.4 & 47.8
& 78.3 & 76.6 & 54.9 \\

Log-prob.
& 31.1 & 44.8 & 36.7
& 61.6 & 78.7 & \textbf{41.3}
& 72.2 & 84.2 & 48.4
& 78.3 & 76.3 & 54.7 \\

Predictive Entropy
& 29.0 & 49.8 & 36.7
& 58.4 & 82.1 & 40.3
& 72.5 & 84.2 & 49.4
& 72.7 & 86.2 & 54.4 \\

MSP
& 32.1 & 42.4 & 36.5
& 61.6 & 78.7 & \textbf{41.3}
& 72.2 & 84.2 & 48.4
& 78.3 & 76.3 & 54.7 \\

Semantic Entropy
& 29.7 & 47.4 & 36.5
& 68.1 & 66.8 & 36.9
& 75.0 & 77.8 & 42.3
& 73.5 & 82.0 & 52.4 \\

Verbalized conf.
& 29.2 & 48.4 & 36.4
& 80.7 & 31.3 & 25.7
& 78.6 & 66.3 & 39.7
& 78.3 & 76.3 & 54.7 \\

\midrule
\multicolumn{13}{l}{\textit{Steering}} \\

Logit bias
& 31.2 & 43.3 & 36.3
& 77.9 & 46.3 & 37.5
& 78.6 & 70.9 & 43.5
& 76.1 & 59.3 & 39.6 \\

Dense steering
& 33.1 & 45.2 & 38.2
& 80.3 & 36.8 & 25.9
& 75.5 & 73.1 & 40.0
& 75.7 & 79.6 & 54.8 \\

Dr.~No (ours)
& 31.5 & 50.3 & \textbf{38.7}
& 80.4 & 40.9 & 30.2
& 77.9 & 77.4 & 45.5
& 75.1 & 84.6 & \textbf{57.6} \\

Dr.~No (ours) + abstention prompt
& 4.0 & 95.8 & 7.6
& 75.1 & 60.2 & 39.5
& 73.1 & 92.1 & \textbf{55.5}
& 62.5 & 97.6 & 52.5 \\

\bottomrule
\end{tabular}%
}
\vspace{-1.1em}
\end{table*}

\subsection{Baselines}
\label{sec:baselines}

We compare Dr.~No with prompting and post-hoc uncertainty-based rejection baselines. All baselines leave the model activations unchanged.

\paragraph{No steering and abstention prompting.}
The no-steering baseline evaluates the base model without intervention. The abstention-prompt baseline adds an instruction encouraging the model to refuse when the visual or textual evidence is insufficient, inspired by prior work on eliciting uncertainty and abstention through prompting~\cite{canllmsexpress,dollmsknow}. Full abstention prompt is reported in ~\ref{app:prompting_baseline}.

\paragraph{Token-level uncertainty.}
We compute confidence statistics from a single greedy decoding pass: mean generated-token log-probability, predictive entropy, and maximum softmax probability (MSP). Low log-probability, high entropy, or low MSP are treated as evidence for answer rejection \cite{msp,malinin}.

\paragraph{Self-assessment.}
For $P(\mathrm{True})$, the model is shown its own answer in a second pass and asked whether it is correct; lower probability assigned to \emph{True} indicates lower confidence \cite{kadavath2022language}. For verbalized confidence, the model directly reports a confidence score between 0 and 100 \cite{canllmsexpress}.

\paragraph{Semantic entropy.}
Semantic entropy estimates uncertainty from disagreement across sampled generations by clustering equivalent answers and computing entropy over the resulting answer clusters \cite{kuhn2023semantic}.

\paragraph{Threshold selection.}
For post-hoc rejection methods, thresholds are selected on a validation split
disjoint from the test set and then fixed for test evaluation. The validation
source varies by benchmark: UNK-VQA provides its own split; MM-UPD has none, so
we tune on the unsolvable-problem data released with CLIP-UP, which provides
dedicated train/validation splits~\cite{clipup}; and AbstentionBench has none, so
we tune on SUM (the synthetic corpus used for feature mining), held out from its
evaluation subsets.

\subsection{Models and intervention setup}
\label{sec:models_setup}

\paragraph{Models.}
For the multimodal experiments, we use LLaVA~1.6 Mistral 7B~\cite{llava} with the corresponding LLaVA SAE release~\cite{llava_sae}, mining features from the GQA-based multimodal corpus described in Section~\ref{sec:mining_corpora}. For the Qwen experiments, we use the vision-language Qwen3.5-2B model~\cite{qwen3.5} with the corresponding Qwen SAE release~\cite{qwen_sae}. In the main text, we evaluate Qwen in a language abstention setting on AbstentionBench, mining features from SUM and evaluating on held-out AbstentionBench subsets. Appendix~\ref{app:qwen_vlm_benchmarks} reports Qwen results on the visual abstention benchmarks.

\paragraph{Steering configuration.}
Per benchmark, we select the target layer, number of steered features, steering strength, intervention mode, and gating option on the same validation splits used for the uncertainty baselines, then fix this configuration across all test examples. Table~\ref{tab:drno_config} reports the final configurations; candidate grids, prompts, parsers and other details are provided in the Appendix~\ref{app:implementation_details}.

\subsection{Benchmarks}
\label{sec:benchmarks}

\paragraph{UNK-VQA.}
UNK-VQA evaluates whether a vision-language model can recognise when a visual question is unanswerable from the image~\cite{unkvqa}. Unanswerable examples are constructed by perturbing VQA-style image--question pairs, e.g.\ through noun replacement, attribute modification, or visual-scene mismatch. We use the multiple-choice setting: answerable examples contain one correct option, while unanswerable ones require selecting an explicit \emph{Unanswerable} option.

\paragraph{MM-UPD.}
MM-UPD evaluates unanswerable multiple-choice visual question answering through paired solvable and unsolvable examples \cite{mmupd}. We report results on its three sub-benchmarks. In \textbf{Absent Answer Detection (AAD)}, the image and question are compatible but the correct answer is missing from the candidate set. In \textbf{Incompatible Answer Set Detection (IASD)}, the answer options are incompatible with the question. In \textbf{Incompatible Visual Question Detection (IVQD)}, the question is incompatible with the image. Following the benchmark protocol, we report standard accuracy, UPD accuracy, and dual accuracy, where dual accuracy requires both correctly answering the solvable instance and abstaining on its paired unsolvable variant.

\paragraph{AbstentionBench.}
AbstentionBench evaluates abstention in language reasoning by pairing answerable examples with unanswerable variants \cite{abstentionbench}. We evaluate four subsets: GSM8K-abstain \cite{cobbe2021training}, UMWP \cite{sun2024benchmarking}, MMLU-Math-abstain \cite{hendrycks2021measuring}, and MuSiQue \cite{trivedi2022musique}. These subsets test whether abstention features mined from SUM transfer to mathematical reasoning and retrieval-heavy multi-hop QA.
\subsection{Metrics}
\label{sec:metrics}

\paragraph{Answerable and unanswerable accuracy.}
For UNK-VQA and AbstentionBench, we report accuracy separately on answerable and unanswerable examples. Answerable accuracy measures whether the model selects or generates the correct answer when the input is solvable. Unanswerable accuracy measures whether the model correctly abstains when the input is unsolvable.

Because abstention involves a trade-off between answering and refusing, we use the harmonic mean of answerable and unanswerable accuracy as the primary metric:
\begin{equation}
\mathrm{HM}
=
\frac{
2 \cdot \mathrm{Acc}_{\mathrm{ans}} \cdot \mathrm{Acc}_{\mathrm{unans}}
}{
\mathrm{Acc}_{\mathrm{ans}} + \mathrm{Acc}_{\mathrm{unans}}
}.
\end{equation}
This penalizes methods that improve abstention by collapsing answer accuracy, or vice versa.

\paragraph{MM-UPD metrics.}
For MM-UPD, we follow the benchmark protocol and report standard accuracy, UPD accuracy, and dual accuracy. Standard accuracy measures performance on the original solvable examples. UPD accuracy measures whether the model correctly detects the corresponding unsolvable problem and abstains rather than selecting an invalid answer. Dual accuracy is the primary metric: a paired example is counted as correct only if the model answers the standard solvable instance correctly and also abstains on its corresponding UPD instance. This directly evaluates whether a method preserves useful answering behaviour while detecting the paired unanswerable variant.

\section{Results}
\label{sec:results}

We report results for vision-language abstention with LLaVA~1.6 Mistral 7B (Table~\ref{tab:merged_results}), followed by language reasoning abstention with Qwen3.5-2B (Table~\ref{tab:qwen3_abstentionbench}).
\paragraph{UNK-VQA.}
Table~\ref{tab:merged_results} reports results on the multiple-choice UNK-VQA benchmark. The unsteered model rarely abstains, answering most questions regardless of whether the visual evidence supports a confident response. The post-hoc uncertainty baselines improve the harmonic mean mainly by shifting the model towards more frequent refusal, conceding answerable accuracy in return.

Dr. No achieves the best harmonic mean on UNK-VQA, converting a comparable loss in answering into a larger gain
in abstention than the uncertainty baselines do. The two approaches answer a similar share of solvable questions, but feature steering abstains more reliably on unanswerable ones, reaching a more favourable operating point overall.

\paragraph{MM-UPD.}
Table~\ref{tab:merged_results} shows that Dr.~No behaves differently across the three MM-UPD sub-benchmarks. This contrasts with the clearer trend on UNK-VQA and indicates that MM-UPD probes multiple sources of unanswerability rather than a single abstention skill.

The MM-UPD results support a positive but task-specific interpretation of feature steering. Dr.~No improves over the base model and abstention prompt across the three subsets, with the strongest gains on IASD and IVQD, where abstention depends on detecting answer-set or image-question incompatibility. These settings better match the mismatch patterns in the feature-mining corpus, allowing SAE-based steering to encourage abstention without relying solely on output-level confidence. The main limitation is AAD, where Dr.~No underperforms the strongest post-hoc uncertainty baselines in dual accuracy. Here, the image and question may still imply a plausible latent answer, while the correct option is absent from the candidate set. The task is therefore closer to option-set verification than intrinsic unanswerability of the image--question pair.
In its current form, SAE steering does not fully address this failure mode, suggesting that it may need to be complemented by uncertainty methods tailored to absent-answer cases.

\paragraph{AbstentionBench.}

Table~\ref{tab:qwen3_abstentionbench} evaluates Dr.~No on language reasoning abstention using Qwen3.5-2B across LLM reasoning benchmarks. Averaged over the mathematical reasoning subsets, Dr.~No achieves the strongest harmonic mean, outperforming the unsteered model, abstention prompting, and uncertainty-based rejection baselines. Full, disaggregated results are reported in Appendix~\ref{app:qwen_abstentionbench_full}. The improvement is driven primarily by better recognition of unanswerable problems while largely preserving accuracy on answerable examples, suggesting that feature steering can serve as an effective abstention-control mechanism in language benchmarks as well.

At the same time, the weaker result on MuSiQue shows that this benefit is not uniform across all language tasks: retrieval-heavy multi-hop QA appears to require different or more context-sensitive abstention signals than those sufficient for mathematical reasoning.

\begin{table}[]
\centering
\caption{
Qwen3.5-2B AbstentionBench results.
\textbf{Ans} and \textbf{Unans} denote accuracy on answerable and unanswerable
subsets, respectively.
\textbf{HM} is their harmonic mean.
}
\label{tab:qwen3_abstentionbench}
\setlength{\tabcolsep}{4pt}
\small
\begin{tabular}{lccc}
\toprule
\multicolumn{4}{c}{\textbf{Math average: MMLU-Math, UMWP, GSM8K}} \\
\midrule
\textbf{Method} & \textbf{Ans} & \textbf{Unans} & \textbf{HM} \\
\midrule

\multicolumn{4}{l}{\textit{Baseline}} \\
No steering & 71.3 & 71.0 & 71.1 \\
No steering + abstention prompt & 67.5 & 82.4 & 74.2 \\
\midrule
\multicolumn{4}{l}{\textit{Uncertainty}} \\
P(True) & 68.1 & 84.3 & 75.4 \\
Log-prob. & 69.0 & 84.4 & 75.9 \\
Predictive Entropy & 68.0 & 86.0 & 75.9 \\
MSP & 68.9 & 84.9 & 76.1 \\
Semantic Entropy & 68.5 & 79.9 & 73.8 \\
Verbalized conf. & 69.9 & 72.5 & 71.1 \\
\midrule
\multicolumn{4}{l}{\textit{Steering}} \\
Dr.~No (ours) & 69.2 & 88.2 & \textbf{77.6} \\
Dr.~No + abstention prompt (ours) & 50.7 & 93.2  & 65.7 \\
\midrule
\multicolumn{4}{c}{\textbf{MuSiQue}} \\
\midrule
\textbf{Method} & \textbf{Ans} & \textbf{Unans} & \textbf{HM} \\
\midrule

\multicolumn{4}{l}{\textit{Baseline}} \\
No steering
& 48.3 & 83.3 & \textbf{61.1} \\
No steering + abstention prompt
& 36.0 & 90.9 & 51.6 \\
\midrule

\multicolumn{4}{l}{\textit{Uncertainty}} \\
P(True)
& 47.6 & 84.3 & 60.8 \\
Log-prob.
& 47.1 & 84.0 & 60.4 \\
Predictive Entropy
& 47.8 & 83.7 & 60.8 \\
MSP
& 47.1 & 83.9 & 60.3 \\
Semantic Entropy
& 47.1 & 84.6 & 60.5 \\
Verbalized conf.
& 48.3 & 83.3 & 61.1 \\
\midrule

\multicolumn{4}{l}{\textit{Steering}} \\
Dr.~No (ours)
& 46.0 & 85.7 & 59.9 \\
Dr.~No + abstention prompt (ours) & 36.5 & 86.4  & 51.3 \\
\bottomrule
\end{tabular}
\vspace{-1.45em}
\end{table}

\section{Analysis}
\label{sec:analysis}

The results above show that SAE feature steering can improve abstention, but not uniformly across all sources of unanswerability. We therefore analyse whether the mined feature signal transfers beyond the mining distribution and what kinds of failure modes it captures.

\subsection{Feature separability and transfer}
\label{sec:feature_transfer}
\begin{figure}[t]
\centering
\includegraphics[width=\columnwidth]{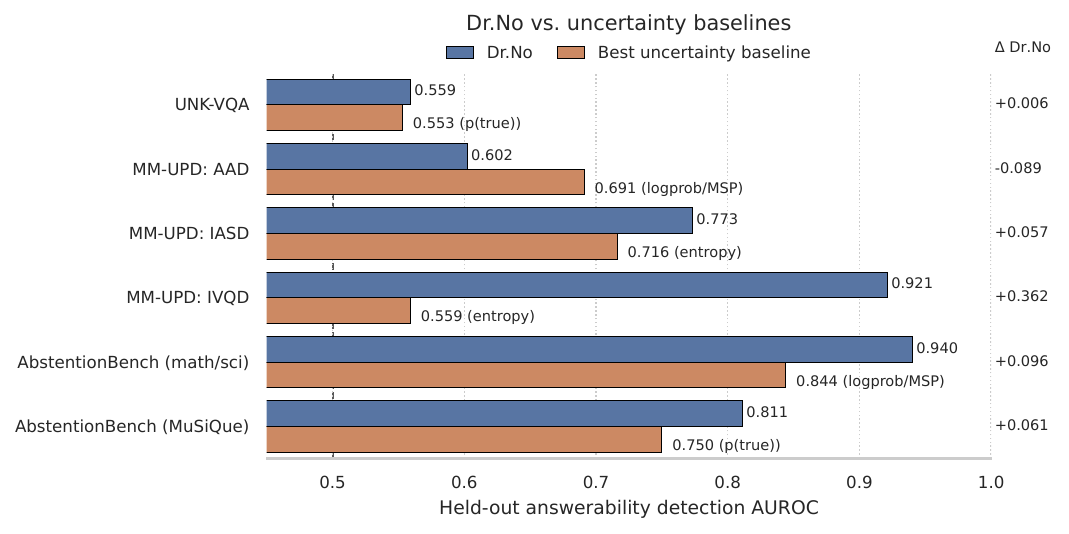}
\caption{\textbf{Answerability detection on held-out evaluation benchmarks.}
Bars report AUROC for distinguishing answerable from unanswerable examples
(higher is better), using features mined on a separate corpus. Dr.~No uses the
mined SAE signal as the detector; the baseline is the best post-hoc uncertainty
method per setting. $\Delta$~Dr.~No is their difference.}
\label{fig:heldout}
\vspace{-1.35em}
\end{figure}

On the GQA mining corpus, individual SAE features strongly separate abstain-labelled examples from commit-labelled, with the signal concentrated in later layers: the strongest feature appears at layer 24, reaching near-perfect separation (Appendix~\ref{app:layer_separability}). This motivates late-layer features as steering targets, though high mining-set separability should not be read as universal answerability detection.

Figure~\ref{fig:heldout} evaluates this transfer by using the mined Dr.~No feature signal as a held-out answerability detector. The signal is competitive on UNK-VQA, strongest on MM-UPD IASD and IVQD, and weakest on AAD. This pattern suggests that the mined features primarily capture incompatibility-driven unanswerability, where the image, question, or answer options are mutually inconsistent. AAD is different: the image and question may still imply a valid latent answer, while only the correct option is missing from the candidate set. This makes AAD closer to option-set verification, which is better captured by output-level uncertainty than by the current steering signal.

The language results show a similar dependence on failure mode. SUM-mined features transfer well to mathematical AbstentionBench subsets, where unanswerability often reflects inconsistent or underspecified problem structure. By contrast, MuSiQue requires retrieval-heavy multi-hop evidence composition, where a simple baseline achieves a better trade-off than our method. The cause is likely representational: decision-token features capture unanswerability present in the static prompt, while MuSiQue's only emerges after multi-hop composition, pointing toward features mined across reasoning steps or steering coupled with evidence verification.

\section{Ethics and Societal Impact}
This work aims to improve reliability by encouraging models to abstain when evidence is insufficient, but the same mechanism is dual-use: it could be inverted to suppress caution or amplified into indiscriminate refusal, including censorship of legitimate answers, with effects likely uneven across inputs that differ from the mining distribution, such as low-resource languages. Our metrics, which penalize indiscriminate refusal and report accuracy on both answerable and unanswerable inputs, rule out the crudest failure mode, but do not establish how abstention is distributed across subpopulations, languages, or visual domains. This differential remains an untested risk, so the method should not be treated as a universal safety or uncertainty detector; before any high-stakes use, abstention behaviour should be audited across subpopulations and content domains and kept under human oversight.

\section{Limitations and Conclusion}

We introduced Dr.~No, an inference-time feature-steering method for improving abstention in frozen foundation models. Dr.~No mines SAE features associated with abstention and answer commitment, then steers a small set of decoder directions without retraining. On UNK-VQA, it achieves a more favourable answer-abstain trade-off than post-hoc uncertainty-based rejection or abstention prompting.


These results show that sparse internal features provide useful control handles for abstention, although their effectiveness depends on the task and failure mode. More broadly, they suggest that the decision to abstain is not governed by a single universal direction but by source-specific signals that respond differently to the same intervention. Future work should mine features from diverse sources of unanswerability, develop adaptive selection and steering strategies, and combine feature-level interventions with calibrated uncertainty for more robust abstention. 

Taken together, our findings suggest that abstention is an interpretable and steerable internal property of foundation models, opening new possibilities for developing systems that exhibit more reliable and adaptive abstention behaviour.
{
    \small
    \bibliographystyle{ieeenat_fullname}
    \bibliography{main}
    \clearpage
}
\appendix
\onecolumn
\section{Implementation details}
\label{app:implementation_details}

\paragraph{Model and SAE checkpoints.}
For the multimodal experiments, we use LLaVA~1.6 Mistral 7B \cite{llava} with the SAE suite (one $8\times$-expansion SAE per layer, $\dsae=32768$ for $\dmodel=4096$) \cite{llava_sae}. The SAEs are trained on the residual-stream output. For the Qwen experiments, we use Qwen3.5-2B \cite{qwen3.5} with the Qwen-Scope residual SAE checkpoint ($\dsae=32768$) \cite{qwen_sae}. All model and SAE checkpoints are kept frozen.

\paragraph{Activation extraction and intervention.}
For feature mining and for the gate readout, we extract residual-stream activations at the \emph{final} prompt token, after the model has processed the image/context, question, answer options, and abstention option. SAE features are computed using the corresponding SAE encoder at the target layer. During inference, steering is applied at the target layer ($\ell^\star\!=\!16$ or $24$ for LLaVA, $15$ for Qwen) to \textbf{all prompt-token positions} during the prefill pass only (the forward pre-hook adds the steering vector to every position with sequence length $>1$). Generated-token activations are not modified. Decoder directions are $\ell_2$-normalised; the steering vector is scaled by the strength $\alpha$ reported in Table~\ref{tab:drno_config}.

Image-derived tokens are treated identically to text tokens: they are part of the prefill sequence, so the same additive steering vector is added to the image-token positions as to the text positions. We do not single out or mask image tokens.

\paragraph{Final steering configurations.}
The final configurations used in the main experiments are reported in Table~\ref{tab:drno_config}. Note that the configuration is benchmark-specific. The two LLaVA benchmarks do not share a single configuration: the reported Dr.~No row for UNK-VQA uses layer~24 with $\alpha=5.0$, whereas MM-UPD uses layer~16 with $\alpha=2.0$ (both use $k=3$, bidirectional mode, and gating). For Qwen, the main-text AbstentionBench transfer uses the single configuration in the Qwen-VL row of Table~\ref{tab:drno_config} (layer~15, $k=3$, $\alpha=1.0$, bidirectional, gated), while the Qwen-VL multimodal results reported in Appendix~\ref{app:qwen_vlm_benchmarks} again use their own per-benchmark configurations. The candidate hyperparameter grid was, for LLaVA, layers $\in\{16,20,24\}$, $k\in\{1,3,5\}$, $\alpha\in\{1.0,2.0,3.0,5.0\}$, mode $\in\{$abstention-up, commitment-down, bidirectional$\}$, and gate $\in\{$off, on\}; and for Qwen, layers $\in\{15,18,21\}$, $k\in\{1,3,5\}$, $\alpha\in\{1.0,2.0,3.0,5.0\}$, the same three modes, and the same gating options. Feature selection (which SAE features enter $\mathcal{A}_k$/$\mathcal{C}_k$) uses a held-out split of the mining corpus; the remaining choices ($\alpha$, mode, gate) are selected on a benchmark validation split, using the same per-benchmark validation sources as the uncertainty baselines: UNK-VQA's own validation split, the CLIP-UP unsolvable-problem train/validation splits for MM-UPD, and SUM for the Qwen AbstentionBench transfer. All selection splits are disjoint from the test sets, and the configuration is fixed before test evaluation.

\paragraph{Decoding.}
Unless otherwise stated, all single-generation evaluations use greedy decoding. We generate at most \textbf{64} new tokens for the multiple-choice benchmarks and \textbf{1024} new tokens for AbstentionBench. For semantic entropy, we sample \textbf{5} responses per input using temperature \textbf{1.0}, top-p \textbf{0.9}, and the same max-new-tokens budget as the corresponding benchmark (64 for multiple-choice, 1024 for AbstentionBench).

\paragraph{Prompt formatting.}
For multiple-choice visual benchmarks, each prompt contains the image, question, answer options, and an explicit abstention option. In the GQA mining corpus and in our UNK-VQA multiple-choice rendering, the option order is randomised per example: the \emph{Unanswerable} option is assigned to a uniformly sampled slot among A--D, and the three answer candidates are placed in the remaining slots. For answerable GQA mining examples, the gold answer is therefore also assigned to a randomly determined non-abstention slot; for unanswerable mining examples, the sampled \emph{Unanswerable} slot is used as the gold label. We store the resulting option-to-label mapping for each example and use that mapping for answer parsing and scoring. The corresponding prompt template is:

\begin{quote}\small\ttfamily

Choose the correct answer from the following options:\\
A) {a}\\
B) {b}\\
C) {c}\\
D) Unanswerable\\
Reply with only the letter (A, B, C, or D).
\end{quote}
For MM-UPD, we follow the benchmark's native prompt format and evaluation protocol. In the option setting, the benchmark appends the UPD-specific abstention option: `None of the above'' for AAD/IASD and `The image and question are irrelevant'' for IVQD, together with the instruction ``Answer with the option's letter from the given choices directly.'' In the instruction setting, we use the benchmark's explicit refusal instruction, which refers to the appended abstention option as slot~F. MM-UPD is evaluated with the benchmark's CircularEval protocol, so each example is assessed under circularly shifted option orders and is counted as correct only according to the benchmark's native circular-evaluation criterion. Thus, the MM-UPD abstention option is represented and scored according to the benchmark-provided option label in each evaluated prompt, rather than by our GQA/UNK-VQA option template. The separate abstention-prompt baseline is described in Appendix~\ref{app:prompting_baseline}; unless that baseline is explicitly enabled, we do not add any additional refusal-oriented system prompt beyond the benchmark's native prompt. Together with the answerable-accuracy, harmonic-mean, and MM-UPD dual-accuracy metrics in Section~\ref{sec:metrics}, this setup reduces the risk that improvements reflect a fixed abstention slot, a single abstention phrase, or a generic refusal bias rather than evidence-sensitive abstention. We add additional analysis with two paraphrase options in section \ref{app:label-paraphrase}. 

\paragraph{Answer parsing.}
For multiple-choice benchmarks, outputs are mapped to answer options using the \textbf{MM-UPD multiple-choice parser} (first-letter / option-string matching), with the optional GPT judge fallback disabled. A prediction is counted as abstention if it selects the explicit abstention option or matches one of the strings \{\textbf{``Unanswerable'', ``None of the above'', ``No answer'', ``I cannot answer''}\}. For AbstentionBench, both answerable and unanswerable examples are scored by the \textbf{official AbstentionBench LLM judge} Llama 3.1 8B Instruct \cite{llama} run locally using the abstention-verifier prompt from \cite{abstentionbench}: an answerable example is correct when the judge matches the response to a reference answer, and an unanswerable example is correct when the judge labels the response an abstention. Judge-indeterminate outputs are dropped from the denominator; ambiguous or unparsable multiple-choice outputs are treated as \textbf{non-abstentions} (i.e.\ incorrect on unanswerable items).

\paragraph{Uncertainty baselines.}
For token-level confidence baselines, we compute scores over the generated answer tokens only. For $P(\mathrm{True})$, the model is shown its own answer and asked to judge whether it is correct using the prompt:
\begin{quote}\small\ttfamily
\{prompt\}\\
\\
Proposed answer: \{answer\}\\
Is the proposed answer correct? Answer True or False.
\end{quote}
and we score the probability assigned to the \emph{True} token. For verbalized confidence, we strip the ``reply with only the letter'' instruction and use:
\begin{quote}\small\ttfamily
\{prompt\}\\
\\
Proposed answer: \{answer\}\\
\\
How confident are you that this answer is correct? Respond with exactly one line, 'Confidence: N', where N is an integer from 0 (pure guess) to 100 (certain).
\end{quote}
For all post-hoc rejection baselines, thresholds are selected on the \textbf{validation split} to maximise the benchmark's primary metric (\textbf{harmonic mean} for UNK-VQA and AbstentionBench, \textbf{mean dual accuracy} for MM-UPD) and then fixed for test evaluation. As described in Section~\ref{sec:baselines}, the validation source is UNK-VQA's own validation split, the CLIP-UP unsolvable-problem train/validation splits for MM-UPD (which has no native validation split), and SUM for AbstentionBench (held out from its evaluation subsets); the Dr.~No steering configuration is selected on these same splits.
\subsection{Mining corpus details}
\label{app:mining_details}

\paragraph{Multimodal mining corpus.}
The multimodal mining corpus is constructed from the GQA balanced training split \cite{gqa}. We draw \textbf{5{,}000} answerable image--question examples and, applying one perturbation of each construction type per source question, build \textbf{20{,}000} synthetic unanswerable examples: \textbf{5{,}000} semantic-drift, \textbf{5{,}000} object-absence, \textbf{5{,}000} false-premise, and \textbf{5{,}000} answer-dropping examples. This yields \textbf{25{,}000} mining items in total. Candidate answer options are generated by sampling distractors from the GQA answer vocabulary: object-absence, false-premise, and semantic-drift items reuse the same candidate-answer layout, while answer-dropping removes the gold option and backfills a distractor. For each mining example, we randomly permute the four option slots A--D, so the \emph{Unanswerable} option is not tied to a fixed position. On answerable examples, the gold answer is placed in one of the four slots according to this sampled permutation; on unanswerable examples, the sampled slot containing \emph{Unanswerable} is used as the gold label. We store the resulting option-to-label mapping for each example and use it for answer parsing, behavioural filtering, and feature mining. The corpus is drawn only from the GQA training split, which is separate from the UNK-VQA and MM-UPD evaluation benchmarks; images and questions are \textbf{not} additionally de-duplicated against the benchmark sources beyond this split-level separation.

\paragraph{Behavioural filtering.}
Because the unsteered model does not always behave correctly on the synthetic mining corpus, we filter examples before feature mining. Abstention-associated features are mined from synthetic unanswerable examples on which the unsteered model correctly selects \emph{Unanswerable}. Commitment-associated features are mined from answerable examples on which the unsteered model selects the correct answer.

\paragraph{Text mining corpus.}
For the text-only transfer experiments, we mine Qwen abstention features from SUM \cite{sum}. Each SUM training problem is unrolled into one answerable and one unanswerable item, yielding \textbf{20000} answerable and \textbf{20000} unanswerable examples. SUM examples are disjoint from the AbstentionBench evaluation subsets.

\section{Computational overhead}
\label{app:latency}
The implementation details above describe how the steering vector is built and applied at inference time; a natural follow-up question is what this adds to ordinary generation. We therefore quantify the runtime and memory cost of gated SAE steering relative to unsteered generation on a frozen 10-item UNK-VQA multiple-choice slice, averaged over 5 timed rounds after a warmup pass.

\begin{table}[h]
\centering

\small
\caption{Overhead of steering relative to unsteered generation.
Wall-clock is reported as mean $\pm$ s.d.\ over 5 rounds.}
\label{tab:latency_overhead}
\setlength{\tabcolsep}{6pt}
\begin{tabular}{lrr}
\toprule
 & \textbf{Baseline} & \textbf{Steered} \\
\midrule
Wall-clock (ms, total)      & $839 \pm 11$ & $847 \pm 12$ \\
Per-item latency (ms)       & 84           & 85           \\
SAE encode cost (ms/round)  & --           & 2.1          \\
Peak GPU memory (GiB)       & 14.8         & 15.3         \\
\bottomrule
\end{tabular}
\end{table}

Relative to unsteered generation, Dr.~No adds \textbf{+8\,ms} to wall-clock
time per round (\textbf{+1.0\%}), of which SAE encoding accounts for
\textbf{2.1\,ms}. Peak GPU memory increases by \textbf{0.5\,GiB}, driven by
the additional \textbf{512\,MiB} of resident SAE decoder/encoder parameters.
Hook registration is a one-time \textbf{2.9\,s} setup cost paid once per
session and excluded from per-example timings. Overall, the intervention adds a small, fixed overhead on top of ordinary generation, dominated by the one-time hook setup rather than by any recurring per-token cost.

\section{Additional Method Details}
\label{app:method_details}

This section expands two components introduced in the main text: the gated
variant of the steering update (Section~\ref{sec:steering}) and the
abstention-prompting baseline (Section~\ref{sec:baselines}). The resulting
mining and steering procedures are summarised as pseudocode in
Appendix~\ref{app:algorithm_pseudocode}.

\subsection{Gated steering}
\label{app:gated_steering}

Uniform steering applies the same intervention strength to every input. While
this can increase abstention on unanswerable examples, it may also introduce a
global refusal bias and unnecessarily reduce accuracy on answerable examples. We
therefore consider a gated variant of Dr.~No in which the intervention strength
is computed from the model's own abstention-related SAE activations for the
current input.

For gated steering, we compute a scalar abstention readout from the SAE
activations of a small set of gate features $\mathcal{G}_k$. By default,
$\mathcal{G}_k$ is chosen as the top-$k$ abstention-associated features mined on
the feature-mining corpus. For an input $x$, the readout is
\begin{equation}
u(x)
=
\frac{1}{|\mathcal{G}_k|}
\sum_{j \in \mathcal{G}_k}
a_j\!\left(h^{(\ell^\star)}_{\mathrm{last}}(x)\right),
\end{equation}
where $h^{(\ell^\star)}_{\mathrm{last}}(x)$ is the hidden state at the target
layer evaluated at the final prompt token.

We map this readout to a gate value via a sigmoid function,
\begin{equation}
g(x)
=
\sigma\!\left(\frac{u(x)-\tau}{T}\right),
\qquad
g(x) \in [0,1],
\end{equation}
where $\tau$ is a threshold and $T>0$ is a temperature, both calibrated on the
mining corpus. The threshold $\tau$ centres the gate between the abstain and
commit readouts, while $T$ sets how sharply the gate transitions around it. The
resulting gate scales the steering vector:
\begin{equation}
\tilde{h}^{(\ell^\star)}_t
=
h^{(\ell^\star)}_t
+
g(x)v(m,\alpha).
\end{equation}

Let $\bar{u}_{\mathrm{abstain}}$ and $\bar{u}_{\mathrm{commit}}$ denote the
average readout values on abstain-labelled and commit-labelled mining examples,
respectively. We set
\begin{equation}
\tau
=
\frac{1}{2}
\left(
\bar{u}_{\mathrm{abstain}}
+
\bar{u}_{\mathrm{commit}}
\right),
\qquad
T
=
\frac{1}{4}
\left(
\bar{u}_{\mathrm{abstain}}
-
\bar{u}_{\mathrm{commit}}
\right).
\end{equation}
Setting $\tau$ to the midpoint of the two class readouts makes the gate larger
when the current input activates the mined abstention features and smaller when
those features are weakly active, while tying $T$ to the abstain--commit
separation rescales the transition to the spread of the readout, so the gate
neither saturates nor stays flat across the operating range observed during
mining.

We also allow an optional gate floor,
\begin{equation}
g'(x)
=
g_{\min}
+
(1-g_{\min})g(x),
\end{equation}
where $g_{\min}=0$ fully suppresses steering when abstention evidence is weak,
and $g_{\min}>0$ preserves a small uniform component of the intervention. When
a gate floor is used, $g'(x)$ replaces $g(x)$ in the steering update. With left padding, the final prompt token corresponds to the decision position
for every example in the batch. The gate is computed once per input during the
prefill pass and broadcast across all prompt-token positions. Activations at
generated tokens are not modified.
\subsection{Abstention prompting baseline}
\label{app:prompting_baseline}

Prompting is a natural baseline because instruction framing can affect whether a
model expresses uncertainty, reports confidence, or refuses to answer
\cite{canllmsexpress,dollmsknow}. Related self-evaluation work also shows that
models can sometimes assess the reliability of their own answers when explicitly
prompted to do so \cite{kadavath2022language}. In abstention settings, however,
the relevant question is not simply whether prompting changes behaviour, but
whether it improves the balance between correct answering and appropriate
refusal.

For the multimodal experiments (UNK-VQA and MM-UPD), the abstention-prompt
baseline supplies the following instruction as the system prompt, leaving the
multiple-choice prompt itself unchanged:
\begin{quote}\small\ttfamily
You are a visual question answering assistant. Your task is to answer questions about images. However, if the question cannot be answered based on the image - because a key object is missing, the question is semantically incoherent, or the answer requires knowledge not present in the image - you must respond with exactly: 'unanswerable'. Do not guess or hallucinate an answer.
\end{quote}
For the text-only AbstentionBench experiments the same instruction is used with
``images''/``image'' replaced by ``text'', and the gate-calibrated mining-time
system prompt of Section~\ref{sec:mining_corpora} is the default for the steered
runs (see Appendix~\ref{app:implementation_details}); ``\,+\,abstention prompt''
rows replace it with the explicit refusal instruction above. The same prompt text
is used for both the unsteered (``No steering + abstention prompt'') and steered
(``Dr.~No + abstention prompt'') rows, so the only difference between those pairs
is the SAE intervention.

This balance matters because UNK-VQA evaluates answerable and unanswerable
examples jointly, while MM-UPD further requires success on paired solvable and
unsolvable cases through dual accuracy. In practice, abstention-oriented
prompting acts as a coarse global control. It encourages more refusals, which
can improve performance on unanswerable examples, but it also increases
unnecessary abstention on answerable ones.

Table~\ref{tab:merged_results} illustrates this pattern. On UNK-VQA, the prompt
substantially increases unanswerable accuracy, but answerable accuracy collapses,
yielding a low harmonic mean. On MM-UPD, the prompt improves average dual
accuracy, but the gain appears to come from a broad shift towards caution rather
than selective abstention.

This motivates feature-based steering. Whereas prompting modifies behaviour
through an instruction-level prior, SAE-based steering aims to intervene more
locally by modulating abstention-related internal features.
\section{Layer-wise Feature Separability}
\label{app:layer_separability}

Table~\ref{tab:top_features_app} reports the strongest abstention-associated SAE
feature at each inspected layer on the GQA mining corpus. For each layer,
features are scored at the decision position by how well their activations
separate abstain-labelled examples from commit-labelled examples.

The results show that separability becomes increasingly concentrated in later
layers. Layer 16 already contains an abstention-relevant feature, but the
strongest inspected signal appears at layer 24, where a single feature reaches
near-perfect separation on the mining corpus.

These values should be interpreted as a feature-mining diagnostic rather than as
held-out answerability detection. High separability on the mining corpus
identifies plausible intervention directions, but it does not imply that the same
single feature detects every form of unanswerability across downstream
benchmarks.

\begin{table}[t]
\centering
\caption{Layer-wise AUROC separability of the strongest abstention-associated
SAE feature on the GQA mining corpus for LLaVA-1.6 Mistral 7B. Mean activations
are reported as abstain / commit.}
\label{tab:top_features_app}
\setlength{\tabcolsep}{6pt}
\small
\begin{tabular}{ccc}
\toprule
\textbf{Layer} & \textbf{AUROC} & \textbf{Mean activation} \\
\midrule
16          & 0.872          & 3.38 / 0.48           \\
20          & 0.928          & 6.24 / 0.12           \\
\textbf{24} & \textbf{0.998} & \textbf{19.87} / 4.53 \\
\bottomrule
\end{tabular}
\end{table}
\section{Algorithm Pseudocode}
\label{app:algorithm_pseudocode}

\begin{algorithm}[h]
\caption{Mining abstention and commitment SAE features}
\label{alg:feature-mining}
\DontPrintSemicolon
\KwIn{
Mining dataset $D_{\mathrm{mine}}$; model; SAE encoder $E$; layer $L$;
number of features top\_k
}
\KwOut{
Abstention feature set $\mathcal{A}$; commitment feature set $\mathcal{C}$; per-feature class means
}

$Z_{\mathrm{abs}}, Z_{\mathrm{com}} \leftarrow [\,], [\,]$\;

\ForEach{$x \in D_{\mathrm{mine}}$}{
  $h \leftarrow$ residual-stream activation at layer $L$, decision position (final prompt token)\;
  $z \leftarrow E(h)$\;

  \eIf{$x$ belongs to the abstention / unanswerable class}{
    append $z$ to $Z_{\mathrm{abs}}$\;
  }{
    append $z$ to $Z_{\mathrm{com}}$\;
  }
}

\ForEach{SAE latent $j$}{
  $\Delta_j \leftarrow
  \operatorname{mean}(Z_{\mathrm{abs}}[:,j])
  -
  \operatorname{mean}(Z_{\mathrm{com}}[:,j])$\;

  store $\operatorname{mean}(Z_{\mathrm{abs}}[:,j])$ and
  $\operatorname{mean}(Z_{\mathrm{com}}[:,j])$\;
  optionally log Cohen's $d$ and AUROC for diagnostics\;
}

$\mathcal{A} \leftarrow$ top-$k$ latents by largest positive $\Delta_j$ (abstention-associated)\;
$\mathcal{C} \leftarrow$ top-$k$ latents by largest negative $\Delta_j$ (commitment-associated)\;

apply the consistency filter for the corresponding mining setting\;
\Return $\mathcal{A}$, $\mathcal{C}$, and class means\;
\end{algorithm}

\begin{algorithm}[h]
\caption{Dr.~No SAE steering with optional gate}
\label{alg:sae-steering}
\DontPrintSemicolon
\KwIn{
SAE decoder $W_{\mathrm{dec}}$; top-$k$ abstention features $\mathcal{A}_k$ and
commitment features $\mathcal{C}_k$; target layer $\ell^\star$; strength $\alpha$;
mode $m$ (abstention-up, commitment-down, or bidirectional); gating flag
}
\KwOut{Steered residual stream at layer $\ell^\star$}

\tcp{Build the steering direction once, before decoding (Eq.~\eqref{eq:steering_vector})}
$v_A \leftarrow$ sum of the unit-normalised decoder columns of $\mathcal{A}_k$\;
$v_C \leftarrow$ sum of the unit-normalised decoder columns of $\mathcal{C}_k$\;
\uIf{$m$ is abstention-up}{
  $v \leftarrow v_A$\;
}
\uElseIf{$m$ is commitment-down}{
  $v \leftarrow -\,v_C$\;
}
\Else{
  $v \leftarrow v_A - v_C$ \tcp*{bidirectional}
}

\BlankLine
\tcp{Steer only on the prefill pass, at every prompt position}
\ForEach{prefill forward pass at layer $\ell^\star$}{
  $h \leftarrow$ residual-stream activations\;
  \eIf{gating is on}{
    $u \leftarrow$ mean activation of the abstention features $\mathcal{A}_k$ at the final prompt token\;
    $g \leftarrow \sigma\!\big((u-\tau)/T\big)$ \tcp*{gate in $[0,1]$; $\tau,T$ from App.~\ref{app:gated_steering}}
  }{
    $g \leftarrow 1$ \tcp*{ungated: full strength}
  }
  $h \leftarrow h + \alpha\,g\,v$\;
}
\Return $h$\;
\end{algorithm}

\section{Per-subset AbstentionBench math results for Qwen3.5-2B}
\label{app:qwen_abstentionbench_full}

Table~\ref{tab:qwen3_full_appendix} reports the full per-subset
AbstentionBench results for Qwen3.5-2B. These results complement the averaged
math results in Table~\ref{tab:qwen3_abstentionbench}. Dr.~No achieves the best
harmonic mean on all three math subsets, with the largest gains coming from
improved abstention on unanswerable examples. The per-subset breakdown shows
that these gains concentrate where the unsteered model already abstains least,
suggesting that steering primarily corrects existing overconfidence rather than
uniformly inflating refusal across all subsets.

\begin{table*}[t]
\centering
\caption{
Full per-subset Qwen3.5-2B AbstentionBench results.
\textbf{Ans} denotes strict accuracy on answerable examples, \textbf{Unans}
denotes recall on unanswerable examples, and \textbf{HM} is their harmonic mean.
Best HM per subset is bolded.
}
\label{tab:qwen3_full_appendix}

\scriptsize
\scalebox{1.4}{%

\begin{tabular}{lccccccccc}
\toprule
& \multicolumn{3}{c}{\textbf{MMLU-Math}}
& \multicolumn{3}{c}{\textbf{UMWP}}
& \multicolumn{3}{c}{\textbf{GSM8K}} \\
\cmidrule(lr){2-4}
\cmidrule(lr){5-7}
\cmidrule(lr){8-10}
\textbf{Method}
& \textbf{Ans} & \textbf{Unans} & \textbf{HM}
& \textbf{Ans} & \textbf{Unans} & \textbf{HM}
& \textbf{Ans} & \textbf{Unans} & \textbf{HM} \\
\midrule
\multicolumn{10}{l}{\textit{Baseline}} \\
No steering
& 57.3 & 68.9 & 62.6
& 80.9 & 54.2 & 64.9
& 75.7 & 89.9 & 82.2 \\
No steering + abstention prompt
& 54.5 & 81.6 & 65.4
& 75.8 & 68.0 & 71.6
& 72.1 & 97.6 & 82.9 \\
\midrule
\multicolumn{10}{l}{\textit{Uncertainty}} \\
P(True)
& 57.3 & 81.6 & 67.3
& 71.8 & 79.8 & 75.6
& 75.3 & 91.6 & 82.6 \\
Log-prob.
& 56.4 & 80.6 & 66.3
& 75.7 & 78.9 & 77.3
& 75.0 & 93.6 & 83.3 \\
Predictive Entropy
& 55.5 & 82.5 & 66.3
& 73.5 & 81.6 & 77.4
& 74.9 & 93.8 & 83.3 \\
MSP
& 56.4 & 81.6 & 66.7
& 75.3 & 79.3 & 77.2
& 75.0 & 93.7 & 83.3 \\
Semantic Entropy
& 57.1 & 63.9 & 60.3
& 73.5 & 84.1 & \textbf{78.4}
& 75.0 & 91.8 & 82.5 \\
Verbalized conf.
& 57.3 & 68.9 & 62.6
& 76.6 & 58.6 & 66.4
& 75.7 & 89.9 & 82.2 \\

\midrule
\multicolumn{10}{l}{\textit{Steering}} \\
Dr.~No (ours)
& 56.4 & 88.4 & \textbf{68.8}
& 74.3 & 80.2 & 77.1
& 77.0 & 95.9 & \textbf{85.5} \\
Dr.~No + abstention prompt (ours)
& 53.4 & 88.4 & 66.6
& 44.0 & 91.3 & 59.4
& 54.7 & 99.8 & 70.7 \\
\bottomrule
\end{tabular}%
}

\end{table*}

\section{Qwen3.5-2B on multimodal abstention benchmarks}
\label{app:qwen_vlm_benchmarks}

We also evaluate Qwen3.5-2B on the multiple-choice UNK-VQA and MM-UPD benchmarks.
Table~\ref{tab:qwen35_merged_vlm_results}
reports UNK-VQA and MM-UPD results in the same merged format used for the main
LLaVA~1.6 Mistral 7B results. Despite the smaller and architecturally distinct
backbone, Dr.~No reproduces the qualitative pattern observed on LLaVA in Table~\ref{tab:merged_results}, remaining
strongest on UNK-VQA and the incompatibility-driven subsets while AAD stays the hardest case for steering. This consistency suggests that absent-answer detection is a backbone-agnostic failure mode that feature steering alone does not resolve, at least in our current set up.

\begin{table*}[t]
\centering
\caption{
Results on UNK-VQA and MM-UPD for Qwen3.5-2B.
\textbf{Ans}/\textbf{Unans} denote answerable and unanswerable accuracy, and
\textbf{HM} denotes their harmonic mean. \textbf{Std}/\textbf{Upd} denote
standard and Unsolvable Problem Detection (UPD) accuracy, and \textbf{Dual}
denotes paired standard--UPD accuracy. All values are percentages. For uncertainty
methods, thresholds are selected to maximise mean dual accuracy per method on
MM-UPD.
}
\label{tab:qwen35_merged_vlm_results}
\scriptsize
\scalebox{1.2}{%
\begin{tabular}{lcccccccccccc}
\toprule
&
\multicolumn{3}{c}{\textbf{UNK-VQA}} &
\multicolumn{3}{c}{\textbf{MM-UPD (AAD)}} &
\multicolumn{3}{c}{\textbf{MM-UPD (IASD)}} &
\multicolumn{3}{c}{\textbf{MM-UPD (IVQD)}} \\
\cmidrule(lr){2-4}
\cmidrule(lr){5-7}
\cmidrule(lr){8-10}
\cmidrule(lr){11-13}
\textbf{Method}
& \textbf{Ans} & \textbf{Unans} & \textbf{HM}
& \textbf{Std} & \textbf{Upd} & \textbf{Dual}
& \textbf{Std} & \textbf{Upd} & \textbf{Dual}
& \textbf{Std} & \textbf{Upd} & \textbf{Dual} \\
\midrule

\multicolumn{13}{l}{\textit{Baseline}} \\
No steering
& 43.4 & 12.6 & 19.5
& 88.3 & 46.9 & 37.7
& 85.5 & 65.1 & 41.4
& 86.8 & 44.2 & 31.2 \\

No steering + abstention prompt
& 43.0 & 16.4 & 23.7
& 87.9 & 49.9 & 38.9
& 85.3 & 82.6 & 55.6
& 86.0 & 63.0 & 44.4 \\

\midrule
\multicolumn{13}{l}{\textit{Uncertainty-based rejection}} \\
P(True)
& 30.8 & 50.7 & 38.3
& 73.9 & 87.0 & \textbf{54.9}
& 70.8 & 91.5 & 49.5
& 76.9 & 85.2 & 55.1 \\

Log-prob.
& 29.0 & 49.4 & 36.5
& 76.1 & 83.4 & 51.1
& 75.0 & 94.7 & 56.8
& 77.8 & 88.2 & 54.5 \\

Predictive Entropy
& 31.2 & 45.8 & 37.1
& 73.7 & 87.5 & 54.2
& 72.1 & 96.8 & 56.9
& 74.1 & 93.3 & \textbf{55.9} \\

MSP
& 29.0 & 49.4 & 36.5
& 76.2 & 83.1 & 51.6
& 75.0 & 94.6 & 56.7
& 77.7 & 88.2 & 54.5 \\

Semantic Entropy
& 26.9 & 53.5 & 35.8
& 83.6 & 63.7 & 41.1
& 79.5 & 85.1 & 47.1
& 81.8 & 65.5 & 34.3 \\

Verbalized conf.
& 37.7 & 28.1 & 32.2
& 88.1 & 47.1 & 37.6
& 85.4 & 65.2 & 41.6
& 86.7 & 44.4 & 31.7 \\

\midrule
\multicolumn{13}{l}{\textit{Steering}} \\

Dr.~No (ours)
& 32.4 & 50.9 & \textbf{39.6}
& 82.3 & 70.7 & 52.2
& 83.8 & 86.5 & 58.5
& 85.7 & 73.6 & 50.7 \\
Dr.~No + abstention prompt (ours)
& 30.2 & 51.5 & 38.1
& 80.6 & 72.1 & 52.5
& 84.0 & 87.4 & \textbf{60.0}
& 84.4 & 77.4 & 54.4 \\
\bottomrule
\end{tabular}%
}
\end{table*}

\section{Robustness Analysis}
\label{app:robustness}

The UNK-VQA and MM-UPD numbers reported in the preceding sections, and in Table~\ref{tab:merged_results}, are point estimates on finite test and mining samples. Before turning to the trade-off and ablation analyses below, we quantify how much of the observed gap between Dr.~No and the baselines is attributable to sampling noise, as opposed to the stochastic choices made during feature mining. We assess robustness along two independent axes: (i) uncertainty in the evaluation itself, arising from finite test-set sampling, and (ii) uncertainty in the feature-mining pipeline that underlies steering, arising from the stochastic choices made during mining. These two sources of variation are estimated with different procedures and should not be conflated.

\subsection{Evaluation Uncertainty (Test-Set Bootstrap)}
\label{app:eval-uncertainty}

To quantify sampling noise in our evaluation metrics, we perform 10,000 paired bootstrap resamples (seed~$=2026$) over the common evaluation examples shared by all methods. Example IDs are normalized across methods before resampling. Each of the 10,000 draws samples the shared example set with replacement and recomputes the harmonic mean (HM) on UNK-VQA, or the mean dual score on MM-UPD, for every method on the identical index vector, so comparisons across methods remain paired. We report the 2.5th and 97.5th percentiles of the resulting distribution as the 95\% confidence interval (CI).

On UNK-VQA, response-steering with a neutral prompt (DRNO~+~neutral) achieves the strongest HM at 38.7 (95\% CI: [38.00, 39.46]), compared with 35.2 ([34.59, 35.84]) for the unsteered neutral baseline. Switching to an abstention prompt sharply changes the answerable/unanswerable trade-off for both steered and unsteered models, at the cost of overall HM.

\begin{table}[h]
\centering

\caption{UNK-VQA paired bootstrap (10,000 resamples).}
\label{tab:unkvqa-bootstrap}
\begin{tabular}{lrrrr}
\toprule
Method & Answerable & Unanswerable & HM & 95\% CI \\
\midrule
No steer + neutral            & 37.5 & 33.2 & 35.2 & [34.59, 35.84] \\
No steer + abstention prompt  & 6.1  & 93.8 & 11.5 & [10.27, 11.89] \\
DRNO + neutral                & 31.5 & 50.3 & \textbf{38.7} & [38.00, 39.46] \\
DRNO + abstention prompt      & 4.0  & 95.8 & 7.6  & [5.17, 8.03] \\
Log probability               & 31.1 & 44.8 & 36.7 & [36.20, 38.27] \\
Predictive entropy            & 29.0 & 49.8 & 36.7 & [36.52, 37.57] \\
MSP                           & 32.1 & 42.4 & 36.5 & [35.97, 37.28] \\
P(True)                       & 30.4 & 48.5 & 37.3 & [36.85, 38.77] \\
Verbalized confidence         & 29.2 & 48.4 & 36.4 & [35.91, 37.16] \\
Semantic entropy              & 29.7 & 47.4 & 36.5 & [36.14, 37.58] \\
\bottomrule
\end{tabular}
\end{table}

On MM-UPD, the ranking reverses in prompting preference: DRNO with an abstention prompt achieves the strongest mean dual score, 49.2 ([46.48, 51.52]), while DRNO with a neutral prompt obtains 44.4 ([38.35, 46.86]).

\begin{table}[h]
\centering

\caption{MM-UPD paired bootstrap (10,000 resamples).}
\label{tab:mmupd-bootstrap}
\begin{tabular}{lrr}
\toprule
Method & Mean dual & 95\% CI \\
\midrule
No steer + neutral            & 38.4 & [34.66, 39.14] \\
No steer + abstention prompt  & 46.8 & [45.80, 50.44] \\
DRNO + neutral                & 44.4 & [38.35, 46.86] \\
DRNO + abstention prompt      & \textbf{49.2} & [46.48, 51.52] \\
Log probability               & 48.1 & [45.81, 50.44] \\
Predictive entropy            & 48.0 & [45.69, 50.39] \\
MSP                           & 48.1 & [45.81, 50.44] \\
P(True)                       & 47.1 & [44.83, 49.44] \\
Verbalized confidence         & 40.0 & [37.77, 42.27] \\
Semantic entropy              & 43.9 & [41.55, 46.19] \\
\bottomrule
\end{tabular}
\end{table}

\subsection{Mining-Seed Robustness (Feature-Selection Uncertainty)}
\label{app:mining-seed}

A separate concern is whether the features selected by our mining pipeline are sensitive to arbitrary stochastic choices made during mining. We reran the entire GQA feature-mining stack under three independent seeds, each seed governing dataset construction (mining-item order, synthetic-unanswerable construction, and multiple-choice option permutations) and the subsequent activation extraction. From each seed's activations we reselected the top-5 response-contrast features independently at layers 16 and 24, then froze the paper's steering recipe on each seed's features and scored the real evaluation sets. This procedure quantifies variability in the feature-mining pipeline itself, not sampling noise in the test examples, and is therefore reported separately from the bootstrap CIs in Section~\ref{app:eval-uncertainty}.

\begin{table}[h]
\centering

\caption{Downstream metrics by mining seed.}
\label{tab:mining-seed}
\begin{tabular}{lrr}
\toprule
Seed & UNK-VQA HM & MM-UPD mean dual \\
\midrule
0 & 38.52 & 49.07\\
1 & 38.73 & 49.10 \\
2 & 38.74 & 49.06 \\
\midrule
Mean $\pm$ SD & $38.70 \pm 0.13$ & $49.08 \pm 0.02$ \\
\bottomrule
\end{tabular}
\end{table}

Downstream results are highly stable across mining seeds: UNK-VQA HM varies by only 0.13 SD (range: 38.52--38.74), and MM-UPD mean dual varies by only 0.02 SD (range: 49.06--49.10). This variation is an order of magnitude smaller than the test-set bootstrap CIs in Tables~\ref{tab:unkvqa-bootstrap}--\ref{tab:mmupd-bootstrap}, indicating that mining-pipeline stochasticity contributes negligibly to the reported performance relative to evaluation-set sampling noise.

\subsection{Feature-Selection Stability}
\label{app:feature-stability}

Finally, we check whether the identities and rankings of the selected features themselves are stable across mining seeds. All four feature families we use for steering yield \textbf{identical top-5 feature sets} across all three seeds.

\begin{table}[h]
\centering

\caption{Top-5 feature selections by mining seed.}
\label{tab:feature-ids}
\begin{tabular}{clrl}
\toprule
Layer & Feature family & Seed & Ranked feature IDs \\
\midrule
16 & Abstention & 0 & 31464, 427, 15608, 5922, 21208 \\
16 & Abstention & 1 & 427, 31464, 15608, 5922, 21208 \\
16 & Abstention & 2 & 427, 31464, 15608, 5922, 21208 \\
16 & Commitment & 0 & 28090, 9135, 991, 15186, 7460 \\
16 & Commitment & 1 & 28090, 9135, 991, 15186, 7460 \\
16 & Commitment & 2 & 28090, 9135, 991, 15186, 7460 \\
24 & Abstention & 0 & 24636, 2421, 20775, 3310, 10096 \\
24 & Abstention & 1 & 24636, 2421, 20775, 3310, 10096 \\
24 & Abstention & 2 & 24636, 2421, 20775, 3310, 10096 \\
24 & Commitment & 0 & 26982, 6360, 7362, 21244, 7796 \\
24 & Commitment & 1 & 26982, 6360, 21244, 7362, 7796 \\
24 & Commitment & 2 & 26982, 6360, 21244, 7362, 7796 \\
\bottomrule
\end{tabular}
\end{table}

Rank ordering is nearly as stable as set membership. Across the three pairwise seed comparisons for each feature family, the mean Spearman rank correlation is 1.000 for two of the four families (layer-16 commitment and layer-24 abstention, which show no rank changes at all) and 0.933 for the other two (layer-16 abstention and layer-24 commitment), where the only discrepancy in each case is a single swap in rank between two of the same five features (e.g., features 31464 and 427 exchange the first two ranks at layer 16 for abstention).

\begin{table}[h]
\centering

\caption{Pairwise overlap and rank stability across mining seeds.}
\label{tab:rank-stability}
\begin{tabular}{clrrr}
\toprule
Layer & Feature family & Seed pairs compared & Intersection & Mean rank correlation \\
\midrule
16 & Abstention & 3 & 5/5 & 0.933 \\
16 & Commitment & 3 & 5/5 & 1.000 \\
24 & Abstention & 3 & 5/5 & 1.000 \\
24 & Commitment & 3 & 5/5 & 0.933 \\
\bottomrule
\end{tabular}
\end{table}

We include the complete confidence-interval tables, seed-level results, feature IDs, and implementation details above for full transparency; further implementation notes will accompany the released code and configuration files.

\section{Pareto trade-off}
\label{app:pareto}
\begin{figure*}[t]
\centering
\begin{minipage}[b]{0.56\textwidth}
  \centering
  \small
  \setlength{\tabcolsep}{4pt}
  \begin{tabular}{lcc}
  \toprule
  \textbf{Variant} & \textbf{UNK-VQA HM} & \textbf{MM-UPD avg. Dual} \\
  \midrule
  Final Dr.~No & 38.7 & 44.4 \\
  Random SAE features & 35.6 & 38.1 \\
  Abstention-up only & 38.2 & 42.2 \\
  Commitment-down only & 36.3 & 39.7 \\
  \bottomrule
  \end{tabular}
  \captionof{table}{
  Ablations for Dr.~No on representative LLaVA multimodal metrics.
  UNK-VQA is evaluated by harmonic mean (HM), and MM-UPD by average dual
  accuracy across AAD, IASD, and IVQD. Each variant changes one component
  relative to the final configuration.
  }
  \label{tab:drno_ablations}
\end{minipage}
\hfill
\begin{minipage}[b]{0.40\textwidth}
  \centering
  \includegraphics[width=\linewidth]{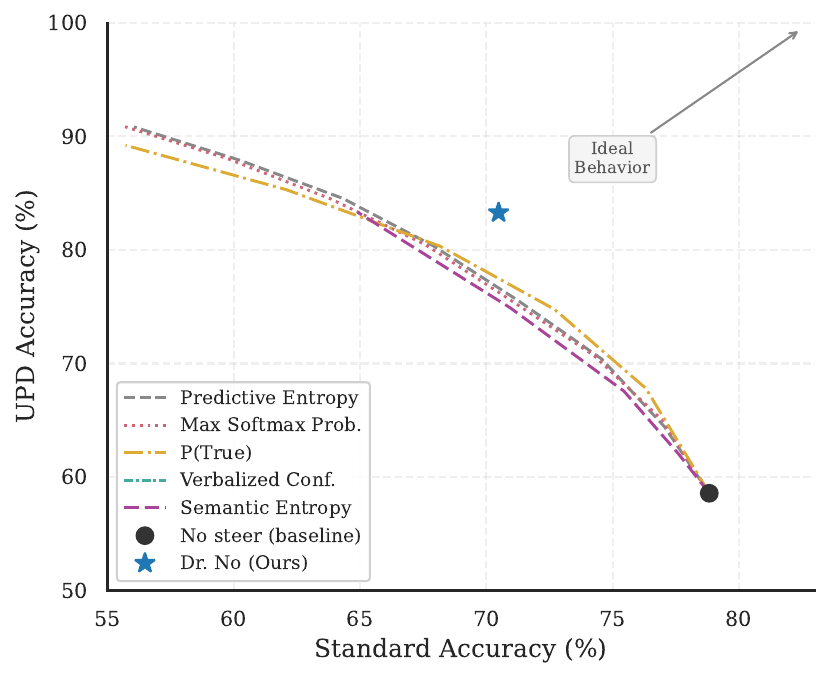}
  \captionof{figure}{Pareto trade-off between answering and abstention
  performance on MM-UPD.}
  \label{fig:pareto}
\end{minipage}
\end{figure*}

Ideally, a model would attain perfect standard accuracy on answerable examples and perfect abstention accuracy on unanswerable examples, corresponding to the upper-right corner of Figure~\ref{fig:pareto}. In practice, however, model behaviour is constrained by imperfect optimization, noisy training data, limited training budgets, model capacity, and other approximation error. Abstention should therefore be analyzed as a Pareto problem: a useful method should improve abstention on unanswerable examples without unnecessarily sacrificing accuracy on answerable ones. Figure~\ref{fig:pareto} visualizes this trade-off across uncertainty-estimation methods. Most post-hoc rejection baselines move along a common frontier, raising UPD accuracy only by reducing standard accuracy. In contrast, Dr.~No shifts the attainable trade-off outward, achieving a more favourable balance between answer preservation and appropriate abstention. Although feature steering is not universal and remains sensitive to the source of unanswerability, these results suggest that SAE-derived interventions improve abstention more selectively than thresholding output-level uncertainty scores alone.

\section{Component ablations}
\label{app:ablations}

Table~\ref{tab:drno_ablations} ablates the main components of Dr.~No on the
LLaVA multimodal benchmarks, changing one factor at a time relative to the final
configuration. Replacing the mined SAE features with randomly selected features
sharply degrades both UNK-VQA harmonic mean and MM-UPD dual accuracy, confirming
that the gains stem from the specific mined directions rather than from generic
activation perturbation. Among the steering modes, the bidirectional
configuration used by Dr.~No is strongest: abstention-up alone retains most of
the benefit, whereas commitment-down alone is markedly weaker. This indicates
that increasing abstention-associated activity is the more important of the two
effects, while suppressing commitment directions contributes a smaller
complementary gain.

\section{Dense Steering Baseline}
\label{sec:dense_steering}

The ablation above isolates the contribution of the mined SAE features and steering mode \emph{within} the sparse-dictionary approach. A complementary question is whether the sparse dictionary itself is necessary, or whether an equivalent intervention applied directly in the raw residual stream achieves similar gains. To evaluate model steering in the absence of a learned sparse dictionary, we implement a \textbf{dense} steering baseline. Dense steering adopts the exact same steering formulation as SAE steering, but modifies the latent activations directly in the residual stream of the base neural network rather than operating in the sparse feature space of an SAE.

All other experimental variables remain identical to the main setup. Specifically, the feature mining process, prompt templates, evaluation splits, and target intervention points are kept unchanged to ensure a fair comparison.
Across all settings, layer $L = 24$ with steering strength $\alpha = 3$ achieves the best configuration, attaining both the highest UNK-VQA harmonic mean ($38.2\%$) and the highest mean MM-UPD dual accuracy ($40.3\%$), still short of Dr.~No's SAE-based results in Table~\ref{tab:drno_ablations}.

\begin{table*}[t]
\centering

\caption{
Effect of layer $L$ and steering strength $\alpha$ on UNK-VQA and MM-UPD for dense steering with LLaVA~1.6 Mistral 7B. \textbf{Ans}/\textbf{Unans} denote answerable and unanswerable accuracy, and \textbf{HM} denotes their harmonic mean. \textbf{Std}/\textbf{Upd} denote standard and Unsolvable Problem Detection (UPD) accuracy, and \textbf{Dual} denotes paired standard-UPD accuracy. All values are percentages.
}
\label{tab:llava_layer_alpha_results}
\scriptsize
\scalebox{1.2}{%
\begin{tabular}{cccccccccccccc}
\toprule
& &
\multicolumn{3}{c}{\textbf{UNK-VQA}} &
\multicolumn{3}{c}{\textbf{MM-UPD (AAD)}} &
\multicolumn{3}{c}{\textbf{MM-UPD (IASD)}} &
\multicolumn{3}{c}{\textbf{MM-UPD (IVQD)}} \\
\cmidrule(lr){3-5}
\cmidrule(lr){6-8}
\cmidrule(lr){9-11}
\cmidrule(lr){12-14}
\textbf{$L$} & \textbf{$\alpha$}
& \textbf{Ans} & \textbf{Unans} & \textbf{HM}
& \textbf{Std} & \textbf{Upd} & \textbf{Dual}
& \textbf{Std} & \textbf{Upd} & \textbf{Dual}
& \textbf{Std} & \textbf{Upd} & \textbf{Dual} \\
\midrule
16 & 1 & 22.1 & 65.1 & 33.0 & 68.0 & 54.0 & 34.0 & 65.6 & 68.1 & 42.8 & 42.3 & 85.3 & 38.7 \\
16 & 2 & 14.8 & 82.9 & 25.0 & 61.3 & 71.2 & 29.0 & 56.1 & 75.3 & 37.1 & 40.1 & 89.8 & 31.5 \\
16 & 3 & 1.6  & 95.8 & 3.1  & 44.3 & 90.5 & 24.9 & 43.3 & 78.7 & 27.5 & 27.2 & 99.5 & 18.0 \\
16 & 4 & 0.1  & 99.1 & 0.2  & 12.7 & 90.8 & 11.5 & 21.9 & 90.7 & 15.2 & 13.6 & 100.0 & 13.6 \\
16 & 5 & 0.1  & 99.5 & 0.2  & 2.0  & 98.6 & 1.5  & 2.1  & 99.2 & 2.1  & 0.3  & 100.0 & 0.3  \\
\midrule
24 & 1 & 34.2 & 43.1 & 38.2 & 81.2 & 34.5 & 25.2 & 78.9 & 65.3 & 36.7 & 76.5 & 78.2 & 52.2 \\
24 & 2 & 33.9 & 42.7 & 37.8 & 82.6 & 35.4 & 25.6 & 74.4 & 66.2 & 38.3 & 75.8 & 78.2 & 52.6 \\
24 & 3 & 33.1 & 45.2 & 38.2 & 80.3 & 36.8 & 26.0 & 75.5 & 73.1 & 40.0 & 75.7 & 79.6 & 54.8 \\
24 & 4 & 27.5 & 50.6 & 35.7 & 79.5 & 42.4 & 30.5 & 74.5 & 75.2 & 36.0 & 63.5 & 80.8 & 50.7 \\
24 & 5 & 27.3 & 53.7 & 36.2 & 78.8 & 43.3 & 31.3 & 71.7 & 75.3 & 35.9 & 54.1 & 88.1 & 40.5 \\
\bottomrule
\end{tabular}%
}
\end{table*}

\section{Sweeping steering strength for Dr.~No}
\label{app:alpha_sweep}

Having compared sparse and dense interventions at their respective best configurations above, we now examine more finely how sensitive Dr.~No's SAE-based steering is to the strength hyperparameter $\alpha$ itself, extending the per-benchmark grid search summarised in Table~\ref{tab:drno_config}. Sweeping $\alpha$ from 0 to 7 in Tables~\ref{tab:unk_vqa_alpha_steering} and \ref{tab:mm_upd_alpha_steering} reveals that Dr.~No's optimal steering strength differs across benchmarks. On UNK-VQA, larger $\alpha$ steadily shifts accuracy from answerable to unanswerable questions, with the harmonic mean peaking at $\alpha=5$ (HM = 38.7\%) before declining slightly at higher values.

On MM-UPD, the trade-off is sharper: standard accuracy drops and UPD accuracy rises as $\alpha$ increases, but the Dual metric, which rewards both jointly, peaks earlier at $\alpha=2$. This suggests that MM-UPD is more sensitive to over-steering, favoring a milder intervention than UNK-VQA to avoid sacrificing standard performance for abstention gains.
\begin{table}[t]
\centering

\caption{
Effect of sweeping steering strength ($\alpha$) on UNK-VQA for LLaVA~1.6 Mistral 7B evaluated at layer 24 without abstention prompting. \textbf{Ans} and \textbf{Unans} denote answerable and unanswerable accuracy, and \textbf{HM} denotes their harmonic mean. All values are percentages.
}
\label{tab:unk_vqa_alpha_steering}
\small
\begin{tabular}{cccc}
\toprule
\textbf{$\alpha$} & \textbf{Ans} & \textbf{Unans} & \textbf{HM} \\
\midrule
1.0 & 35.7 & 38.1 & 36.9 \\
2.0 & 34.3 & 41.9 & 37.7 \\
3.0 & 33.4 & 44.8 & 38.3 \\
4.0 & 32.5 & 47.6 & 38.6 \\
5.0 & 31.5 & 50.3 & 38.7 \\
6.0 & 30.3 & 52.3 & 38.4 \\
7.0 & 29.5 & 54.4 & 38.2 \\
\bottomrule
\end{tabular}
\end{table}
\begin{table*}[t]
\centering

\caption{
Effect of sweeping steering strength ($\alpha$) on MM-UPD for LLaVA~1.6 Mistral 7B evaluated at layer 16 with abstention prompting enabled. \textbf{Std} and \textbf{Upd} denote standard and Unsolvable Problem Detection (UPD) accuracy, and \textbf{Dual} denotes paired standard and UPD accuracy. All values are percentages.
}
\label{tab:mm_upd_alpha_steering}
\scriptsize
\scalebox{1.2}{%
\begin{tabular}{cccccccccc}
\toprule
&
\multicolumn{3}{c}{\textbf{MM-UPD (AAD)}} &
\multicolumn{3}{c}{\textbf{MM-UPD (IASD)}} &
\multicolumn{3}{c}{\textbf{MM-UPD (IVQD)}} \\
\cmidrule(lr){2-4}
\cmidrule(lr){5-7}
\cmidrule(lr){8-10}
\textbf{$\alpha$}
& \textbf{Std} & \textbf{Upd} & \textbf{Dual}
& \textbf{Std} & \textbf{Upd} & \textbf{Dual}
& \textbf{Std} & \textbf{Upd} & \textbf{Dual} \\
\midrule
1.0 & 80.9 & 35.7 & 27.1 & 78.6 & 71.7 & 41.9 & 76.2 & 81.6 & 55.9 \\
2.0 & 80.4 & 40.9 & 30.2 & 77.9 & 77.4 & 45.5 & 75.1 & 84.6 & 57.6 \\
3.0 & 78.7 & 41.0 & 28.8 & 78.6 & 73.9 & 44.7 & 75.8 & 82.4 & 56.7 \\
4.0 & 77.2 & 46.9 & 26.4 & 75.6 & 80.1 & 43.2 & 73.8 & 86.5 & 53.0 \\
5.0 & 73.9 & 48.4 & 25.8 & 76.1 & 82.5 & 38.7 & 74.3 & 90.1 & 52.2 \\
6.0 & 70.2 & 54.1 & 22.5 & 72.9 & 86.9 & 37.2 & 70.9 & 91.8 & 46.8 \\
7.0 & 62.9 & 55.7 & 21.0 & 61.3 & 87.7 & 32.5 & 61.8 & 94.9 & 44.2 \\
\bottomrule
\end{tabular}%
}
\end{table*}

\section{Logit Bias Analysis}
\label{sec:appendix_logit_bias}

As a simpler, dictionary-free point of comparison to sweeping $\alpha$ in the residual stream, we also consider intervening directly at the output layer: applying a logit bias by adding a scalar bias value to the unnormalized logit of the abstention option prior to greedy selection. This shifts predictions toward abstention whenever the top content-answer logit exceeds the abstention option's initial logit by less than the applied bias.

Table~\ref{tab:logit_bias_results} reports the impact of varying the logit bias from $0.5$ to $10.0$ across UNK-VQA and MM-UPD benchmarks. Increasing the bias steadily improves unanswerable and unsolvable problem detection accuracy across all benchmarks, reaching near-perfect detection at higher values. However, aggressive biasing introduces a tradeoff against standard answerable accuracy. Balanced performance peaks at intermediate bias ranges (around $1.0$ to $4.0$), where the harmonic mean on UNK-VQA and dual accuracy metrics across MM-UPD subsets achieve their optimal trade-offs before excessive abstention degrades overall utility.

\section{Abstention-Label Paraphrase Robustness}
\label{app:label-paraphrase}

The strength and logit-bias sweeps above vary \emph{how hard} the model is pushed toward abstention; a separate question is whether that push depends on the specific \emph{wording} used to express the abstention option. To verify that steering modulates internal representations of visual sufficiency rather than surface-level token preferences, we evaluate Dr.~No on UNK-VQA under semantically equivalent paraphrases of the unanswerable option. All prompts, evaluation instances, model parameters, and steering hyperparameters remain identical. Performance is reported across answerable accuracy (Ans), unanswerable accuracy (Unans), and their harmonic mean (HM).
\begin{table*}[]
\centering

\caption{
Effect of logit bias on UNK-VQA and MM-UPD for LLaVA~1.6 Mistral 7B. \textbf{Ans}/\textbf{Unans} denote answerable and unanswerable accuracy, and \textbf{HM} denotes their harmonic mean. \textbf{Std}/\textbf{Upd} denote standard and Unsolvable Problem Detection (UPD) accuracy, and \textbf{Dual} denotes paired standard--UPD accuracy. All values are percentages.
}
\label{tab:logit_bias_results}
\scriptsize
\scalebox{1.2}{%
\begin{tabular}{ccccccccccccc}
\toprule
&
\multicolumn{3}{c}{\textbf{UNK-VQA}} &
\multicolumn{3}{c}{\textbf{MM-UPD (AAD)}} &
\multicolumn{3}{c}{\textbf{MM-UPD (IASD)}} &
\multicolumn{3}{c}{\textbf{MM-UPD (IVQD)}} \\
\cmidrule(lr){2-4}
\cmidrule(lr){5-7}
\cmidrule(lr){8-10}
\cmidrule(lr){11-13}
\textbf{Bias}
& \textbf{Ans} & \textbf{Unans} & \textbf{HM}
& \textbf{Std} & \textbf{Upd} & \textbf{Dual}
& \textbf{Std} & \textbf{Upd} & \textbf{Dual}
& \textbf{Std} & \textbf{Upd} & \textbf{Dual} \\
\midrule
0.5 & 32.8 & 37.1 & 34.8 & 81.1 & 35.0 & 25.9 & 78.8 & 66.1 & 37.7 & 76.1 & 56.3 & 36.0 \\
1.0 & 31.2 & 43.3 & 36.3 & 78.0 & 46.3 & 37.5 & 78.6 & 71.0 & 43.5 & 76.1 & 59.3 & 39.6 \\
2.0 & 29.4 & 53.8 & 38.0 & 72.5 & 49.9 & 33.3 & 64.2 & 75.0 & 36.1 & 75.7 & 63.4 & 40.3 \\
3.0 & 23.7 & 64.9 & 34.7 & 63.9 & 58.7 & 35.0 & 63.6 & 79.8 & 39.3 & 75.3 & 66.5 & 42.4 \\
4.0 & 16.5 & 77.1 & 27.2 & 58.6 & 72.2 & 37.6 & 60.2 & 84.3 & 41.5 & 73.3 & 72.5 & 45.1 \\
5.0 &  9.2 & 88.4 & 16.6 & 51.5 & 85.2 & 32.4 & 60.1 & 87.7 & 42.5 & 70.6 & 77.3 & 45.5 \\
7.0 &  0.8 & 99.2 &  1.6 & 47.2 & 98.9 & 30.2 & 54.9 & 95.0 & 38.6 & 57.1 & 89.0 & 39.1 \\
10.0 &  0.0 & 100.0 &  0.0 & 31.3 & 100.0 & 16.2 & 30.0 & 99.9 & 16.1 & 21.8 & 98.8 & 12.4 \\
\bottomrule
\end{tabular}%
}
\end{table*}

\begin{table}[h]
\centering

\small
\caption{UNK-VQA performance under abstention-label paraphrases.}
\label{tab:label_paraphrase}
\setlength{\tabcolsep}{6pt}
\begin{tabular}{lcccccc}
\toprule
& \multicolumn{3}{c}{\textbf{Baseline}} & \multicolumn{3}{c}{\textbf{Dr.~No}} \\
\cmidrule(lr){2-4} \cmidrule(lr){5-7}
\textbf{Abstention Phrasing} & \textbf{Ans} & \textbf{Unans} & \textbf{HM} & \textbf{Ans} & \textbf{Unans} & \textbf{HM} \\
\midrule
Unanswerable & 37.5 & 33.2 & 35.2 & 31.5 & 50.3 & \textbf{38.7} \\
The image does not provide enough information & 32.9 & 46.1 & 34.8 & 25.6 & 63.0 & \textbf{38.4} \\
No listed answer is supported by the image & 37.9 & 32.7 & 35.1 & 31.0 & 50.3 & \textbf{38.4} \\
\bottomrule
\end{tabular}
\end{table}

As shown in Table~\ref{tab:label_paraphrase}, Dr.~No consistently improves HM by 3.3 to 3.6 points across all phrasing variants by substantially elevating unanswerable detection while preserving balanced answerable performance. These uniform improvements confirm that the mined features govern semantic abstention under insufficient visual evidence rather than an idiosyncratic preference for the literal string ``Unanswerable''.

\section{Qualitative example}
\begin{figure*}[t]
\centering

\begin{tabular}{cc}
\includegraphics[width=0.46\linewidth]{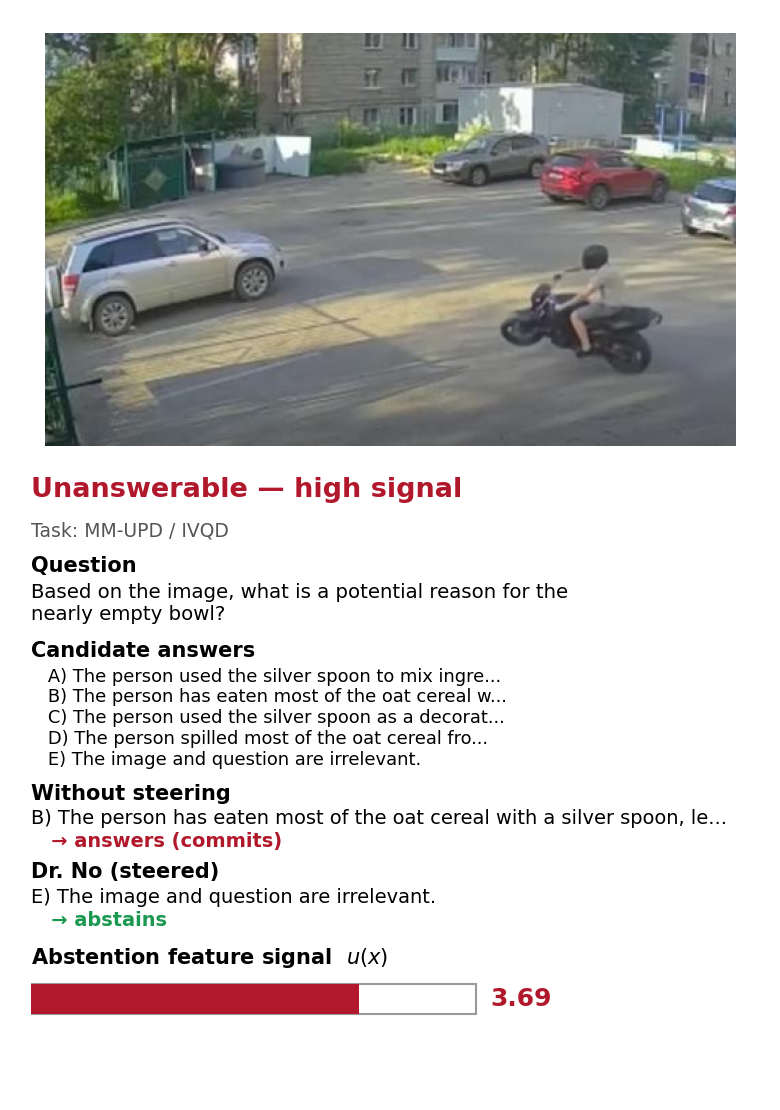} &
\includegraphics[width=0.46\linewidth]{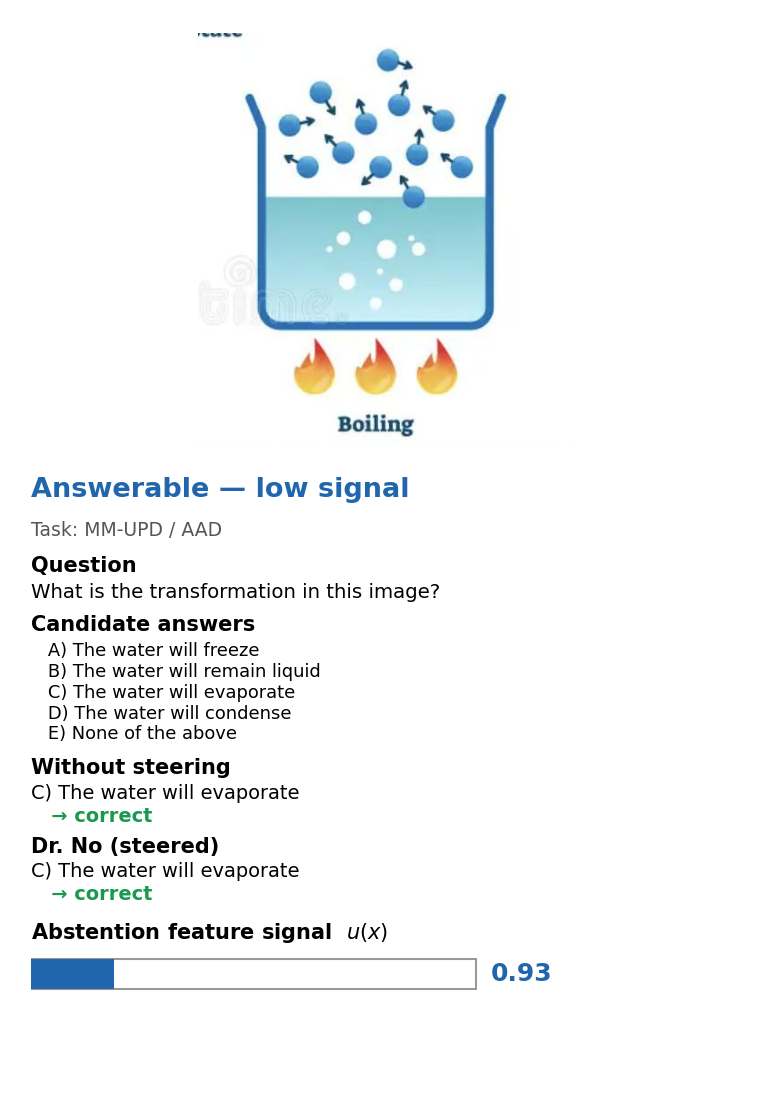} \\
\end{tabular}

\caption{
Effect of Dr.~No steering versus no steering on two MM-UPD inputs. Each panel reports the image, question, candidate answers, the abstention feature signal $u(x)$, and the model's selected option \emph{with} and \emph{without} steering. On the unanswerable example (left) the abstention signal from Dr.~No features
is high: the unsteered model commits to an incorrect
option, whereas Dr.~No abstains. On the answerable example (right) the signal is
low and the steered output is unchanged from the
correct unsteered answer.}
\label{fig:steered_examples}
\end{figure*}

The examples in Figure~\ref{fig:steered_examples} provide an instance-level view of the same feature signal reported in Table~\ref{tab:top_features_app}. In the unanswerable MM-UPD example, the abstention readout is high and steering changes an incorrect committed answer into an abstention; in the answerable example, the readout is low and the steered output remains unchanged. This qualitative behaviour is consistent with the AUROC and mean-activation gaps in Table~\ref{tab:top_features_app}: the mined features can act as useful intervention signals when the input resembles the incompatibility patterns captured during mining. At the same time, these examples should be read as illustrative rather than exhaustive, since the held-out results in Figure~\ref{fig:heldout} show that the same signal transfers unevenly across different sources of unanswerability.

\end{document}

%% file: main.bib
@String(CVPR= {IEEE Conf. Comput. Vis. Pattern Recog.})

@String(ICLR = {Int. Conf. Learn. Represent.})

@String(CVPR  = {CVPR})

@String(ICLR  = {ICLR})

@article{kadavath2022language,
  title={Language models (mostly) know what they know},
  author={Kadavath, Saurav and Conerly, Tom and Askell, Amanda and Henighan, Tom and Drain, Dawn and Perez, Ethan and Schiefer, Nicholas and Hatfield-Dodds, Zac and DasSarma, Nova and Tran-Johnson, Eli and others},
  journal={arXiv preprint arXiv:2207.05221},
  year={2022}
}

@inproceedings{kuhn2023semantic,
  author       = {Lorenz Kuhn and
                  Yarin Gal and
                  Sebastian Farquhar},
  title        = {Semantic Uncertainty: Linguistic Invariances for Uncertainty Estimation
                  in Natural Language Generation},
  booktitle    = {The Eleventh International Conference on Learning Representations,
                  {ICLR} 2023, Kigali, Rwanda, May 1-5, 2023},
  publisher    = {OpenReview.net},
  year         = {2023},
  url          = {https://openreview.net/forum?id=VD-AYtP0dve},
  bibsource    = {dblp computer science bibliography, https://dblp.org}
}

@inproceedings{canllmsexpress,
  author       = {Miao Xiong and
                  Zhiyuan Hu and
                  Xinyang Lu and
                  Yifei Li and
                  Jie Fu and
                  Junxian He and
                  Bryan Hooi},
  title        = {Can LLMs Express Their Uncertainty? An Empirical Evaluation of Confidence
                  Elicitation in LLMs},
  booktitle    = {The Twelfth International Conference on Learning Representations,
                  {ICLR} 2024, Vienna, Austria, May 7-11, 2024},
  publisher    = {OpenReview.net},
  year         = {2024},
  url          = {https://openreview.net/forum?id=gjeQKFxFpZ},
  bibsource    = {dblp computer science bibliography, https://dblp.org}
}

@inproceedings{msp,
  author       = {Dan Hendrycks and
                  Kevin Gimpel},
  title        = {A Baseline for Detecting Misclassified and Out-of-Distribution Examples
                  in Neural Networks},
  booktitle    = {5th International Conference on Learning Representations, {ICLR} 2017,
                  Toulon, France, April 24-26, 2017, Conference Track Proceedings},
  publisher    = {OpenReview.net},
  year         = {2017},
  url          = {https://openreview.net/forum?id=Hkg4TI9xl},
  bibsource    = {dblp computer science bibliography, https://dblp.org}
}

@inproceedings{malinin,
  author       = {Andrey Malinin and
                  Mark J. F. Gales},
  title        = {Uncertainty Estimation in Autoregressive Structured Prediction},
  booktitle    = {9th International Conference on Learning Representations, {ICLR} 2021,
                  Virtual Event, Austria, May 3-7, 2021},
  publisher    = {OpenReview.net},
  year         = {2021},
  url          = {https://openreview.net/forum?id=jN5y-zb5Q7m},
  bibsource    = {dblp computer science bibliography, https://dblp.org}
}

@inproceedings{dollmsknow,
  author       = {Nishanth Madhusudhan and
                  Sathwik Tejaswi Madhusudhan and
                  Vikas Yadav and
                  Masoud Hashemi},
  editor       = {Owen Rambow and
                  Leo Wanner and
                  Marianna Apidianaki and
                  Hend Al{-}Khalifa and
                  Barbara Di Eugenio and
                  Steven Schockaert},
  title        = {Do LLMs Know When to {NOT} Answer? Investigating Abstention Abilities
                  of Large Language Models},
  booktitle    = {Proceedings of the 31st International Conference on Computational
                  Linguistics, {COLING} 2025, Abu Dhabi, UAE, January 19-24, 2025},
  pages        = {9329--9345},
  publisher    = {Association for Computational Linguistics},
  year         = {2025},
  url          = {https://aclanthology.org/2025.coling-main.627/},
  bibsource    = {dblp computer science bibliography, https://dblp.org}
}

@inproceedings{pope,
  author       = {Yifan Li and
                  Yifan Du and
                  Kun Zhou and
                  Jinpeng Wang and
                  Wayne Xin Zhao and
                  Ji{-}Rong Wen},
  editor       = {Houda Bouamor and
                  Juan Pino and
                  Kalika Bali},
  title        = {Evaluating Object Hallucination in Large Vision-Language Models},
  booktitle    = {Proceedings of the 2023 Conference on Empirical Methods in Natural
                  Language Processing, {EMNLP} 2023, Singapore, December 6-10, 2023},
  pages        = {292--305},
  publisher    = {Association for Computational Linguistics},
  year         = {2023},
  url          = {https://doi.org/10.18653/v1/2023.emnlp-main.20},
  doi          = {10.18653/V1/2023.EMNLP-MAIN.20},
  bibsource    = {dblp computer science bibliography, https://dblp.org}
}

@inproceedings{abstentionbench,
  author       = {Polina Kirichenko and
                  Mark Ibrahim and
                  Kamalika Chaudhuri and
                  Samuel J. Bell},
  editor       = {Danielle Belgrave and
                  Cheng Zhang and
                  Laura N. Montoya and
                  Hsuan{-}Tien Lin and
                  Razvan Pascanu and
                  Piotr Koniusz and
                  Marzyeh Ghassemi and
                  Nancy Chen and
                  Iv{\'{a}}n Vladimir Meza Ru{\'{\i}}z and
                  Arturo Loaiza{-}Bonilla},
  title        = {AbstentionBench: Reasoning LLMs Fail on Unanswerable Questions},
  booktitle    = {Advances in Neural Information Processing Systems 38: Annual Conference
                  on Neural Information Processing Systems 2025, NeurIPS 2025, San Diago,
                  CA, USA, December 2-7, 2025 / Mexico City, Mexico, November 30 - December
                  5, 2025},
  year         = {2025},
  url          = {http://papers.nips.cc/paper\_files/paper/2025/hash/fb122bfc3f0127a94ded048b5b03496f-Abstract-Datasets\_and\_Benchmarks\_Track.html},
  bibsource    = {dblp computer science bibliography, https://dblp.org}
}

@article{unkvqa,
  author       = {Yangyang Guo and
                  Fangkai Jiao and
                  Zhiqi Shen and
                  Liqiang Nie and
                  Mohan S. Kankanhalli},
  title        = {{UNK-VQA:} {A} Dataset and a Probe Into the Abstention Ability of
                  Multi-Modal Large Models},
  journal      = {{IEEE} Trans. Pattern Anal. Mach. Intell.},
  volume       = {46},
  number       = {12},
  pages        = {10284--10296},
  year         = {2024},
  url          = {https://doi.org/10.1109/TPAMI.2024.3437288},
  doi          = {10.1109/TPAMI.2024.3437288},
  bibsource    = {dblp computer science bibliography, https://dblp.org}
}

@inproceedings{mmupd,
  author       = {Atsuyuki Miyai and
                  Jingkang Yang and
                  Jingyang Zhang and
                  Yifei Ming and
                  Qing Yu and
                  Go Irie and
                  Yixuan Li and
                  Hai Helen Li and
                  Ziwei Liu and
                  Kiyoharu Aizawa},
  editor       = {Wanxiang Che and
                  Joyce Nabende and
                  Ekaterina Shutova and
                  Mohammad Taher Pilehvar},
  title        = {Unsolvable Problem Detection: Robust Understanding Evaluation for
                  Large Multimodal Models},
  booktitle    = {Proceedings of the 63rd Annual Meeting of the Association for Computational
                  Linguistics (Volume 1: Long Papers), {ACL} 2025, Vienna, Austria,
                  July 27 - August 1, 2025},
  pages        = {6497--6540},
  publisher    = {Association for Computational Linguistics},
  year         = {2025},
  url          = {https://doi.org/10.18653/v1/2025.acl-long.326},
  doi          = {10.18653/V1/2025.ACL-LONG.326},
  bibsource    = {dblp computer science bibliography, https://dblp.org}
}

@inproceedings{llava,
  author       = {Haotian Liu and
                  Chunyuan Li and
                  Qingyang Wu and
                  Yong Jae Lee},
  editor       = {Alice Oh and
                  Tristan Naumann and
                  Amir Globerson and
                  Kate Saenko and
                  Moritz Hardt and
                  Sergey Levine},
  title        = {Visual Instruction Tuning},
  booktitle    = {Advances in Neural Information Processing Systems 36: Annual Conference
                  on Neural Information Processing Systems 2023, NeurIPS 2023, New Orleans,
                  LA, USA, December 10 - 16, 2023},
  year         = {2023},
  url          = {http://papers.nips.cc/paper\_files/paper/2023/hash/6dcf277ea32ce3288914faf369fe6de0-Abstract-Conference.html},
  bibsource    = {dblp computer science bibliography, https://dblp.org}
}

@misc{qwen3.5,
    title  = {{Qwen3.5}: Towards Native Multimodal Agents},
    author = {{Qwen Team}},
    year   = {2026},
    month  = {February},
    url    = {https://qwen.ai/blog?id=qwen3.5}
}

@inproceedings{sum,
  author       = {Linxin Song and
                  Taiwei Shi and
                  Jieyu Zhao},
  editor       = {Christos Christodoulopoulos and
                  Tanmoy Chakraborty and
                  Carolyn Rose and
                  Violet Peng},
  title        = {The Hallucination Tax of Reinforcement Finetuning},
  booktitle    = {Findings of the Association for Computational Linguistics: {EMNLP}
                  2025, Suzhou, China, November 4-9, 2025},
  pages        = {2105--2120},
  publisher    = {Association for Computational Linguistics},
  year         = {2025},
  url          = {https://doi.org/10.18653/v1/2025.findings-emnlp.112},
  doi          = {10.18653/V1/2025.FINDINGS-EMNLP.112},
  bibsource    = {dblp computer science bibliography, https://dblp.org}
}

@article{cobbe2021training,
  author       = {Karl Cobbe and
                  Vineet Kosaraju and
                  Mohammad Bavarian and
                  Mark Chen and
                  Heewoo Jun and
                  Lukasz Kaiser and
                  Matthias Plappert and
                  Jerry Tworek and
                  Jacob Hilton and
                  Reiichiro Nakano and
                  Christopher Hesse and
                  John Schulman},
  title        = {Training Verifiers to Solve Math Word Problems},
  journal      = {CoRR},
  volume       = {abs/2110.14168},
  year         = {2021},
  url          = {https://arxiv.org/abs/2110.14168},
  eprinttype   = {arXiv},
  eprint       = {2110.14168},
  bibsource    = {dblp computer science bibliography, https://dblp.org}
}

@inproceedings{sun2024benchmarking,
  author       = {Yuhong Sun and
                  Zhangyue Yin and
                  Qipeng Guo and
                  Jiawen Wu and
                  Xipeng Qiu and
                  Hui Zhao},
  editor       = {Nicoletta Calzolari and
                  Min{-}Yen Kan and
                  V{\'{e}}ronique Hoste and
                  Alessandro Lenci and
                  Sakriani Sakti and
                  Nianwen Xue},
  title        = {Benchmarking Hallucination in Large Language Models Based on Unanswerable
                  Math Word Problem},
  booktitle    = {Proceedings of the 2024 Joint International Conference on Computational
                  Linguistics, Language Resources and Evaluation, {LREC/COLING} 2024,
                  20-25 May, 2024, Torino, Italy},
  pages        = {2178--2188},
  publisher    = {{ELRA} and {ICCL}},
  year         = {2024},
  url          = {https://aclanthology.org/2024.lrec-main.196},
  bibsource    = {dblp computer science bibliography, https://dblp.org}
}

@inproceedings{hendrycks2021measuring,
  author       = {Dan Hendrycks and
                  Collin Burns and
                  Steven Basart and
                  Andy Zou and
                  Mantas Mazeika and
                  Dawn Song and
                  Jacob Steinhardt},
  title        = {Measuring Massive Multitask Language Understanding},
  booktitle    = {9th International Conference on Learning Representations, {ICLR} 2021,
                  Virtual Event, Austria, May 3-7, 2021},
  publisher    = {OpenReview.net},
  year         = {2021},
  url          = {https://openreview.net/forum?id=d7KBjmI3GmQ},
  bibsource    = {dblp computer science bibliography, https://dblp.org}
}

@article{trivedi2022musique,
  author  = {Trivedi, Harsh and Balasubramanian, Niranjan and Khot, Tushar and Sabharwal, Ashish},
  title   = {MuSiQue: Multihop Questions via Single-hop Question Composition},
  journal = {Transactions of the Association for Computational Linguistics},
  volume  = {10},
  pages   = {539--554},
  year    = {2022},
  doi     = {10.1162/TACL_A_00475},
  url     = {https://doi.org/10.1162/tacl_a_00475}
}

@article{zou2023representation,
  author       = {Andy Zou and
                  Long Phan and
                  Sarah Li Chen and
                  James Campbell and
                  Phillip Guo and
                  Richard Ren and
                  Alexander Pan and
                  Xuwang Yin and
                  Mantas Mazeika and
                  Ann{-}Kathrin Dombrowski and
                  Shashwat Goel and
                  Nathaniel Li and
                  Michael J. Byun and
                  Zifan Wang and
                  Alex Mallen and
                  Steven Basart and
                  Sanmi Koyejo and
                  Dawn Song and
                  Matt Fredrikson and
                  J. Zico Kolter and
                  Dan Hendrycks},
  title        = {Representation Engineering: {A} Top-Down Approach to {AI} Transparency},
  journal      = {CoRR},
  volume       = {abs/2310.01405},
  year         = {2023},
  url          = {https://doi.org/10.48550/arXiv.2310.01405},
  doi          = {10.48550/ARXIV.2310.01405},
  eprinttype   = {arXiv},
  eprint       = {2310.01405},
  bibsource    = {dblp computer science bibliography, https://dblp.org}
}

@article{turner2023steering,
  title={Steering language models with activation engineering},
  author={Turner, Alexander Matt and Thiergart, Lisa and Leech, Gavin and Udell, David and Vazquez, Juan J and Mini, Ulisse and MacDiarmid, Monte},
  journal={arXiv preprint arXiv:2308.10248},
  year={2023}
}

@inproceedings{huben,
  author       = {Robert Huben and
                  Hoagy Cunningham and
                  Logan Riggs Smith and
                  Aidan Ewart and
                  Lee Sharkey},
  title        = {Sparse Autoencoders Find Highly Interpretable Features in Language
                  Models},
  booktitle    = {The Twelfth International Conference on Learning Representations,
                  {ICLR} 2024, Vienna, Austria, May 7-11, 2024},
  publisher    = {OpenReview.net},
  year         = {2024},
  url          = {https://openreview.net/forum?id=F76bwRSLeK},
  bibsource    = {dblp computer science bibliography, https://dblp.org}
}

@article{bricken2023monosemanticity,
       title={Towards Monosemanticity: Decomposing Language Models With Dictionary Learning},
       author={Bricken, Trenton and Templeton, Adly and Batson, Joshua and Chen, Brian and Jermyn, Adam and Conerly, Tom and Turner, Nick and Anil, Cem and Denison, Carson and Askell, Amanda and Lasenby, Robert and Wu, Yifan and Kravec, Shauna and Schiefer, Nicholas and Maxwell, Tim and Joseph, Nicholas and Hatfield-Dodds, Zac and Tamkin, Alex and Nguyen, Karina and McLean, Brayden and Burke, Josiah E and Hume, Tristan and Carter, Shan and Henighan, Tom and Olah, Christopher},
       year={2023},
       journal={Transformer Circuits Thread},
       note={https://transformer-circuits.pub/2023/monosemantic-features/index.html}
    }

@inproceedings{gao,
  author       = {Leo Gao and
                  Tom Dupr{\'{e}} la Tour and
                  Henk Tillman and
                  Gabriel Goh and
                  Rajan Troll and
                  Alec Radford and
                  Ilya Sutskever and
                  Jan Leike and
                  Jeffrey Wu},
  title        = {Scaling and evaluating sparse autoencoders},
  booktitle    = {The Thirteenth International Conference on Learning Representations,
                  {ICLR} 2025, Singapore, April 24-28, 2025},
  publisher    = {OpenReview.net},
  year         = {2025},
  url          = {https://openreview.net/forum?id=tcsZt9ZNKD},
  bibsource    = {dblp computer science bibliography, https://dblp.org}
}

@article{sae_pach,
  author       = {Mateusz Pach and
                  Shyamgopal Karthik and
                  Quentin Bouniot and
                  Serge J. Belongie and
                  Zeynep Akata},
  title        = {Sparse Autoencoders Learn Monosemantic Features in Vision-Language
                  Models},
  journal      = {CoRR},
  volume       = {abs/2504.02821},
  year         = {2025},
  url          = {https://doi.org/10.48550/arXiv.2504.02821},
  doi          = {10.48550/ARXIV.2504.02821},
  eprinttype   = {arXiv},
  eprint       = {2504.02821},
  bibsource    = {dblp computer science bibliography, https://dblp.org}
}

@inproceedings{save,
  author       = {Sangha Park and
                  Seungryong Yoo and
                  Jisoo Mok and
                  Sungroh Yoon},
  title        = {{SAVE:} Sparse Autoencoder-Driven Visual Information Enhancement for
                  Mitigating Object Hallucination},
  booktitle    = {{IEEE/CVF} Winter Conference on Applications of Computer Vision, {WACV}
                  2026, Tucson, AZ, USA, March 6-10, 2026},
  pages        = {7935--7944},
  publisher    = {{IEEE}},
  year         = {2026},
  url          = {https://doi.org/10.1109/WACV61042.2026.00766},
  doi          = {10.1109/WACV61042.2026.00766},
  bibsource    = {dblp computer science bibliography, https://dblp.org}
}

@inproceedings{gqa,
  author       = {Drew A. Hudson and
                  Christopher D. Manning},
  title        = {{GQA:} {A} New Dataset for Real-World Visual Reasoning and Compositional
                  Question Answering},
  booktitle    = {{IEEE} Conference on Computer Vision and Pattern Recognition, {CVPR}
                  2019, Long Beach, CA, USA, June 16-20, 2019},
  pages        = {6700--6709},
  publisher    = {Computer Vision Foundation / {IEEE}},
  year         = {2019},
  url          = {http://openaccess.thecvf.com/content\_CVPR\_2019/html/Hudson\_GQA\_A\_New\_Dataset\_for\_Real-World\_Visual\_Reasoning\_and\_Compositional\_CVPR\_2019\_paper.html},
  doi          = {10.1109/CVPR.2019.00686},
  bibsource    = {dblp computer science bibliography, https://dblp.org}
}

@article{arrogance,
  title={Evaluating large language models for accuracy incentivizes hallucinations},
  author={Kalai, Adam Tauman and Nachum, Ofir and Vempala, Santosh S and Zhang, Edwin},
  journal={Nature},
  pages={1--3},
  year={2026},
  publisher={Nature Publishing Group UK London}
}

@article{selective1,
  author       = {C. K. Chow},
  title        = {On optimum recognition error and reject tradeoff},
  journal      = {{IEEE} Trans. Inf. Theory},
  volume       = {16},
  number       = {1},
  pages        = {41--46},
  year         = {1970},
  url          = {https://doi.org/10.1109/TIT.1970.1054406},
  doi          = {10.1109/TIT.1970.1054406},
  bibsource    = {dblp computer science bibliography, https://dblp.org}
}

@article{selective2,
  author       = {Ran El{-}Yaniv and
                  Yair Wiener},
  title        = {On the Foundations of Noise-free Selective Classification},
  journal      = {J. Mach. Learn. Res.},
  volume       = {11},
  pages        = {1605--1641},
  year         = {2010},
  url          = {https://dl.acm.org/doi/10.5555/1756006.1859904},
  doi          = {10.5555/1756006.1859904},
  bibsource    = {dblp computer science bibliography, https://dblp.org}
}

@inproceedings{selective3,
  author       = {Yonatan Geifman and
                  Ran El{-}Yaniv},
  editor       = {Isabelle Guyon and
                  Ulrike von Luxburg and
                  Samy Bengio and
                  Hanna M. Wallach and
                  Rob Fergus and
                  S. V. N. Vishwanathan and
                  Roman Garnett},
  title        = {Selective Classification for Deep Neural Networks},
  booktitle    = {Advances in Neural Information Processing Systems 30: Annual Conference
                  on Neural Information Processing Systems 2017, December 4-9, 2017,
                  Long Beach, CA, {USA}},
  pages        = {4878--4887},
  year         = {2017},
  url          = {https://proceedings.neurips.cc/paper/2017/hash/4a8423d5e91fda00bb7e46540e2b0cf1-Abstract.html},
  bibsource    = {dblp computer science bibliography, https://dblp.org}
}

@article{qwen_sae,
  author       = {Boyi Deng and
                  Xu Wang and
                  Yaoning Wang and
                  Yu Wan and
                  Yubo Ma and
                  Baosong Yang and
                  Haoran Wei and
                  Jialong Tang and
                  Huan Lin and
                  Ruize Gao and
                  Tianhao Li and
                  Qian Cao and
                  Xuancheng Ren and
                  Xiaodong Deng and
                  An Yang and
                  Fei Huang and
                  Dayiheng Liu and
                  Jingren Zhou},
  title        = {Qwen-Scope: Turning Sparse Features into Development Tools for Large
                  Language Models},
  journal      = {CoRR},
  volume       = {abs/2605.11887},
  year         = {2026},
  url          = {https://doi.org/10.48550/arXiv.2605.11887},
  doi          = {10.48550/ARXIV.2605.11887},
  eprinttype   = {arXiv},
  eprint       = {2605.11887},
  bibsource    = {dblp computer science bibliography, https://dblp.org}
}

@article{llava_sae,
  author       = {Achyuta Rajaram and
                  Sarah Schwettmann and
                  Jacob Andreas and
                  Arthur Conmy},
  title        = {Line of Sight: On Linear Representations in VLLMs},
  journal      = {CoRR},
  volume       = {abs/2506.04706},
  year         = {2025},
  url          = {https://doi.org/10.48550/arXiv.2506.04706},
  doi          = {10.48550/ARXIV.2506.04706},
  eprinttype   = {arXiv},
  eprint       = {2506.04706},
  bibsource    = {dblp computer science bibliography, https://dblp.org}
}

@article{llama,
  author       = {Llama Team},
  title        = {The Llama 3 Herd of Models},
  journal      = {CoRR},
  volume       = {abs/2407.21783},
  year         = {2024},
  url          = {https://doi.org/10.48550/arXiv.2407.21783},
  doi          = {10.48550/ARXIV.2407.21783},
  eprinttype   = {arXiv},
  eprint       = {2407.21783},
  bibsource    = {dblp computer science bibliography, https://dblp.org}
}

@inproceedings{clipup,
  author       = {Ben Vardi and
                  Oron Nir and
                  Ariel Shamir},
  title        = {{CLIP-UP:} CLIP-Based Unanswerable Problem Detection for Visual Question
                  Answering},
  booktitle    = {{IEEE/CVF} Winter Conference on Applications of Computer Vision, {WACV}
                  2026, Tucson, AZ, USA, March 6-10, 2026},
  pages        = {5898--5908},
  publisher    = {{IEEE}},
  year         = {2026},
  url          = {https://doi.org/10.1109/WACV61042.2026.00571},
  doi          = {10.1109/WACV61042.2026.00571},
  bibsource    = {dblp computer science bibliography, https://dblp.org}
}

@inproceedings{guo2017,
  author       = {Chuan Guo and
                  Geoff Pleiss and
                  Yu Sun and
                  Kilian Q. Weinberger},
  editor       = {Doina Precup and
                  Yee Whye Teh},
  title        = {On Calibration of Modern Neural Networks},
  booktitle    = {Proceedings of the 34th International Conference on Machine Learning,
                  {ICML} 2017, Sydney, NSW, Australia, 6-11 August 2017},
  series       = {Proceedings of Machine Learning Research},
  volume       = {70},
  pages        = {1321--1330},
  publisher    = {{PMLR}},
  year         = {2017},
  url          = {http://proceedings.mlr.press/v70/guo17a.html},
  bibsource    = {dblp computer science bibliography, https://dblp.org}
}

@inproceedings{inference_eliciting,
  author       = {Kenneth Li and
                  Oam Patel and
                  Fernanda B. Vi{\'{e}}gas and
                  Hanspeter Pfister and
                  Martin Wattenberg},
  editor       = {Alice Oh and
                  Tristan Naumann and
                  Amir Globerson and
                  Kate Saenko and
                  Moritz Hardt and
                  Sergey Levine},
  title        = {Inference-Time Intervention: Eliciting Truthful Answers from a Language
                  Model},
  booktitle    = {Advances in Neural Information Processing Systems 36: Annual Conference
                  on Neural Information Processing Systems 2023, NeurIPS 2023, New Orleans,
                  LA, USA, December 10 - 16, 2023},
  year         = {2023},
  url          = {http://papers.nips.cc/paper\_files/paper/2023/hash/81b8390039b7302c909cb769f8b6cd93-Abstract-Conference.html},
  bibsource    = {dblp computer science bibliography, https://dblp.org}
}

@inproceedings{safety_refusal,
  author       = {Andy Arditi and
                  Oscar Obeso and
                  Aaquib Syed and
                  Daniel Paleka and
                  Nina Panickssery and
                  Wes Gurnee and
                  Neel Nanda},
  editor       = {Amir Globersons and
                  Lester Mackey and
                  Danielle Belgrave and
                  Angela Fan and
                  Ulrich Paquet and
                  Jakub M. Tomczak and
                  Cheng Zhang},
  title        = {Refusal in Language Models Is Mediated by a Single Direction},
  booktitle    = {Advances in Neural Information Processing Systems 37: Annual Conference
                  on Neural Information Processing Systems 2024, NeurIPS 2024, Vancouver,
                  BC, Canada, December 10 - 15, 2024},
  year         = {2024},
  url          = {http://papers.nips.cc/paper\_files/paper/2024/hash/f545448535dfde4f9786555403ab7c49-Abstract-Conference.html},
  bibsource    = {dblp computer science bibliography, https://dblp.org}
}
